\pdfoutput=1
\documentclass[final, 10pt, twocolumn, twoside]{IEEEtran}
\usepackage{mathrsfs}
\usepackage{ifpdf}
\usepackage{cite}
\usepackage{graphicx}
\usepackage{multirow}
\usepackage{amsfonts}
\usepackage{amssymb}
\usepackage{amsthm}
\usepackage{algorithm}
\usepackage{algorithmic}
\usepackage{array}
\usepackage[marginal]{footmisc}
\usepackage[caption=false,font=footnotesize]{subfig}
\usepackage{stfloats}
\usepackage[usenames]{color}
\usepackage{multirow}
\usepackage{url}
\usepackage{amsmath}
\usepackage{booktabs}
\usepackage{enumerate}
\usepackage{gensymb}
\usepackage{bbm}
\usepackage{stmaryrd}
\usepackage{float}
\usepackage{textcomp}
\usepackage{boxedminipage}
\usepackage{extarrows}
\usepackage{array}
\usepackage{bm}
\usepackage[colorlinks,
            linkcolor=black,
            citecolor=black,
            urlcolor=black
            ]{hyperref}
\makeatletter
\newcommand{\thickhline}{%
    \noalign {\ifnum 0=`}\fi \hrule height 1pt
    \futurelet \reserved@a \@xhline
}
\newcolumntype{"}{@{\hskip\tabcolsep\vrule width 1pt\hskip\tabcolsep}}
\makeatother

\begin{document}
%
\title{Ensemble of Part Detectors for Simultaneous Classification and Localization}
\author{{\small Xiaopeng~Zhang,} {\small Hongkai~Xiong,} \IEEEmembership{\small Senior Member,~IEEE,} {\small Weiyao~Lin,} {\small Qi~Tian,}~\IEEEmembership{\small Fellow,~IEEE}

\thanks{\newline \footnotesize X.~Zhang, H.~Xiong, and W.~Lin are with the Department of Electronic Engineering, Shanghai Jiao Tong University, Shanghai 200240, China. E-mail: \{zxphistory, xionghongkai, wylin\}@sjtu.edu.cn.
\newline \footnotesize Q.~Tian is with the Department of Computer Science, University of Texas at San Antonio, Texas, TX 78249. E-mail: qitian@cs.utsa.edu.}
}
\maketitle

\begin{abstract}
Part-based representation has been proven to be effective for a variety of visual applications. However, automatic discovery of discriminative parts without object\,/\,part-level annotations is challenging. This paper proposes a discriminative mid-level representation paradigm based on the responses of a collection of part detectors, which only requires the image-level labels. Towards this goal, we first develop a detector-based spectral clustering method to mine the representative and discriminative mid-level patterns for detector initialization. The advantage of the proposed pattern mining technology is that the distance metric based on detectors only focuses on discriminative details, and a set of such grouped detectors offer an effective way for consistent pattern mining. Relying on the discovered patterns, we further formulate the detector learning process as a confidence-loss sparse Multiple Instance Learning (cls-MIL) task, which considers the diversity of the positive samples, while avoid drifting away the well localized ones by assigning a confidence value to each positive sample. The responses of the learned detectors can form an effective mid-level image representation for both image classification and object localization. Experiments conducted on benchmark datasets demonstrate the superiority of our method over existing approaches.
\end{abstract}


\section{Introduction}
Object parts that capture crucial characteristics of an image are important in a variety of object recognition and related applications. For instance, in Deformable Part Model (DPM) \cite{felzenszwalb2008dpm}, an object is modeled as a set of deformable parts organized in a tree structure. In relative attribute learning \cite{sandeep2014relative}, local parts that are shared across categories are used to learn relative attributes. In fine-grained recognition \cite{yang2012unsupervised}, \cite{zhang2014fused}, distinctive parts such as the head of birds are detected out to enable part-based representation. Nevertheless, obtaining informative parts usually requires object-level \cite{felzenszwalb2008dpm} or even part-level annotations \cite{azizpour2012object}, which is tedious and costly for large-scale datasets. Accordingly, it is desirable to discover these parts with minimal human supervision.

The success of Convolutional Neural Network (CNN) \cite{krizhevsky2012imagenet} has shed light on the possibility of automatically discovering object parts. It has been revealed that \cite{zeiler2014visualizing} the CNN filters at different layers are sensitive to patches with varying receptive fields, \emph{i.e.}, from low-level cues such as the edges and corners in earlier layers to semantically meaningful parts or even the whole object in deeper layers. From the point of detection, the output of the convolutional layers can be interpreted as detection scores of multiple detectors. In this sense, CNN learns detectors relevant for the dataset it is trained from. However, since the network is trained based on image-level classification losses, these detectors (the hidden layers) are trained \emph{implicitly}. As a result, the discriminative power of the CNN detectors is rather weak, producing activations with inhomogeneous appearances. Though a collection of such weak detectors boost the representative ability, it still leaves room for improvement by enhancing these weak filters.

\begin{figure}[t]
  \centering
  \includegraphics[width=0.46\textwidth]{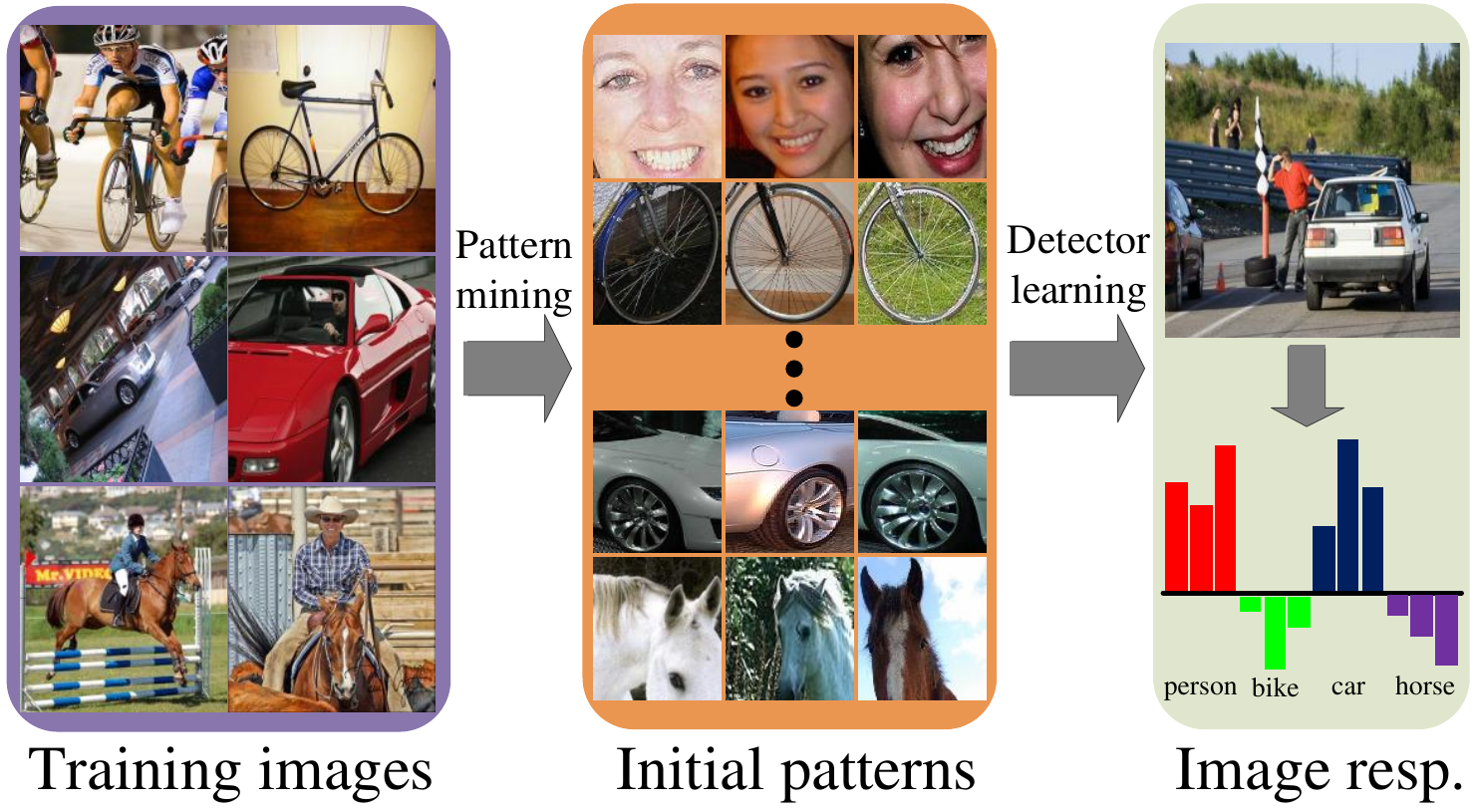}\\
  \vspace{-0.1cm}
  \caption{Image representation based on the part responses. Given a set of training images which are only provided with image-level labels, our goal is to mine mid-level patterns (object parts) that capture crucial aspects of an object, and learn a set of part detectors for image representation.} \label{motivation}
  \vspace{-0.2cm}
\end{figure}

An alternative method of discovering informative parts automatically is to learn detectors \emph{explicitly}, which we refer to weakly supervised detector learning. As shown in Fig. \ref{motivation}, the standard approach for detector learning requires initial patterns (object parts) for detector initialization, and an optimization strategy for detector learning. However, learning part detectors automatically is a classical chicken-and-egg problem: without an accurate appearance model, examples of a part cannot be discovered, while an accurate appearance model cannot be learned without appropriate part exemplars. To solve this challenge, we need to answer the following two crucial issues.

\textbf{What are the right initial patterns?} As the quality of the learned detectors depends heavily on initialization, it is crucial to select appropriate initial patterns. As noted in \cite{singh2012unsupervised}, such patterns should meet two criteria, \emph{i.e.}, representation and discrimination. Representation requires that such patterns should frequently occur in images with the same label, while discrimination claims that they should be seldom found in images not containing the object of interest. Unfortunately, algorithms aim at finding such patterns are rather ad hoc and have limited performance. Most previous works \cite{parizi2014automatic}, \cite{singh2012unsupervised}, \cite{sun2013learning} start from unsupervised clustering such as $k$-means to initialize a part model. However, $k$-means behaves poorly in high dimension since distance metric becomes less meaningful, often producing clustered instances which are inhomogeneous.

\textbf{How to learn generalized detectors?} Given the initial patterns, most weakly supervised learning algorithms follow the pipeline of standard SVM training \cite{malisiewicz2011ensemble}, or an iterative SVM optimization which alternates between training classifier and choosing new positive samples \cite{doersch2013mid}, \cite{juneja2013blocks}, \cite{singh2012unsupervised}. Nevertheless, due to the uncertainty of initial patterns, such optimization is easily to get stuck into a local minimum. On the other hand, due to the occlusion, illumination variation, and viewpoint variation, the same part from different images suffers significant differences. As a result, such methods easily latch on to a few samples which are similar with the initial patterns, but are weak in generalization. Thus, developing optimization strategy under weakly supervised paradigm is important to obtain robust detection performance.


This paper proposes to learn a set of detectors in a weakly supervised paradigm, which aims at solving the above two issues. \textbf{The main contribution} is an iterative optimization strategy for detector learning, which we formulate as a confidence loss sparse Multiple Instance Learning (cls-MIL) task. Different from conventional MIL methods which represents each positive image with a single instance and treats each image equally important, cls-MIL represents each positive image as a sparse linear combination of its member instances, and considers the diversity of the positive images, while avoid drifting away the well localized ones by assigning a confidence value to each positive image. The responses of the learned detectors formulate an effective mid-level image representation for recognition. Another interesting finding is that different from most previous methods which treat image classification \cite{sun2013learning}, \cite{zuo2014learning} and object localization \cite{bilen2015weakly}, \cite{cinbis2014multi}, \cite{li2016image} separately, the proposed approach is able to effectively integrate the two tasks into a whole framework. Benefit from the powerful discriminative ability of the learned part detectors, the detector responses by our approach are able to indicate the locations of the objects. Experiments conducted on benchmark datasets demonstrate the superiority of the proposed representation.

As the detector learning procedure heavily relies initial patterns, \textbf{a second novelty} of our approach is the use of a spectral clustering technology for mining consistent and discriminative patterns. To this end, a selection strategy is first utilized to sample discriminative patches of the corresponding category, followed by exemplar-SVM \cite{malisiewicz2011ensemble} detector training for each sampled patch, finally, these exemplar-SVM detectors are grouped via a spectral clustering strategy for pattern mining. Comparing with traditional clustering methods which are conducted on the original patches, the clustered detectors are able to focus on discriminative details, and a set of such grouped detectors offer an effective way for consistent pattern mining. Furthermore, an entropy coverage criterion is utilized to measure the discriminativeness of each cluster, which enables us to greedily select clusters for detector learning, while not worrying about choosing appropriate number of clusters.

%

The rest of this paper is organized as follows. Sec. II reviews related works on weakly supervised detector learning. The details of our proposed detector learning method are elaborated in Sec. III. In Sec. IV, we apply the learned detectors for classification and localization. Experiments and discussions are given in Sec. V. Finally, Sec. VI concludes the paper.
\section{Related Works}
Over the past years there has been a lot of researches aiming at learning part models in an unsupervised or weakly supervised way. Most methods target at improving the two modules: pattern mining technologies for model initialization, and optimization strategies for detector learning. The learned part models offer a promising way for feature representation, which is beneficial for image recognition and other related applications. In the following, we organize the discussions related to part model learning with the above aspects.
\subsection{Pattern Mining Methods}
Since the ground truth annotations are not available in a weakly supervised paradigm, a number of strategies have been proposed to discover the discriminative patches for model initialization. A simple method, taken in \cite{parizi2014automatic}, \cite{singh2012unsupervised}, \cite{sun2013learning}, starts by randomly sampling a large pool of patches, and employs unsupervised clustering to generate initial patterns for detector learning. Such methods are clumsy and most returned clusters are with inhomogeneous appearances. Hence, many pattern mining technologies are developed to offer better initialization. Song \emph{et al.} \cite{song2014weakly} formulate a constrained submodular algorithm to identify discriminative configurations of visual patterns. Wang \emph{et al.} \cite{wang2014weakly} propose to discover these latent parts via a probabilistic latent semantic analysis on the windows of positive samples and further employ these clusters as sub-categories. Li \emph{et al.} \cite{li2015mid} combine the activations of CNN with the association rule mining technique to discover the representative mid-level patterns. Doersch \emph{et al.} \cite{doersch2013mid} formulate part discovery from the perspective of the well-known mean-shift algorithm to maximize the density ratio in the feature space. There is a special case in which we do not need to worry about exemplar alignment, \emph{i.e.}, a training set consisting exactly of one part instance \cite{malisiewicz2011ensemble}. However, training detectors based on a single exemplar is with limited discriminative power, and the number of detectors scales with the training samples, which is tremendous for large-scale datasets.

Different from previous approaches which aim at grouping the original patches, this paper  performs clustering in terms of the corresponding weak detectors, and makes use of the grouped detectors for pattern mining. In order to generate weak detectors, a selection strategy is first utilized to sample discriminative patches, and each patch is associated with a detector via exemplar-SVM training. Though a single exemplar-SVM detector is weak, a collection of such detectors offer relatively satisfactory localization capacity for pattern mining.

\subsection{Optimization for Detector Learning}
Based on these discovered patterns, most methods employ an iterative learning approach to refine the detectors. Juneja \emph{et al.} \cite{juneja2013blocks} employ an LDA accelerated version \cite{hariharan2012discriminative} of the exemplar-SVMs \cite{malisiewicz2011ensemble}, which reduces the training cost substantially comparing with the standard SVM procedure that involves hard negative mining \cite{felzenszwalb2008dpm}. However, the detectors are trained with only one positive instance, which results in limited discriminative powers. Singh \emph{et al.} \cite{singh2012unsupervised} split the training set into two disjoint parts, and a part model is refined via an iterative procedure which alternates between clustering on one dataset and training discriminative classifiers on the other to avoid overfitting. Parizi \emph{et al.} \cite{parizi2014automatic}  propose a jointly training method which optimizes part models and class specific weights iteratively. Sun \emph{et al.} \cite{sun2013learning} propose a latent SVM model to learn detectors, which tends to select the discriminative parts by enforcing group sparsity regularizer. However, these methods suffer from complex jointly optimization, \emph{e.g.}, \cite{parizi2014automatic} takes over five days to train detectors on MIT Indoor-67 \cite{quattoni2009recognizing}.

The majority of related works treat weakly supervised detector learning as a Multiple Instance Learning (MIL) task, in which labels are assigned to bags (sets of patterns), instead of individual patterns. The positive bags are sets of instances containing at least one positive example, while the negative bags are sets of instances which are all negative. MIL is originally introduced to solve a problem in biochemistry \cite{dietterich1997solving}, and a variety of MIL algorithms have been developed over the years. The simplest method is to transform MIL into a standard supervised learning problem by applying the bag's label to all instances in the bag \cite{ray2005supervised}. However, such method assumes that the positive examples are rich in the positive bags. Andrews \emph{et al.} \cite{andrews2002support} present a new formulation of MIL as a max-margin SVM problem. Bunescu \emph{et al.} \cite{bunescu2007multiple} develop an MIL method which is particularly effective when the positive bags are sparse. When applying MIL for detector learning, the detector is obtained by an iterative procedure which alternates between selecting the highest scoring detection per bag as positive instance and refining the detector models \cite{blaschko2010simultaneous}. However, such simplified setting is sensitive to initialization and easy to getting stuck in a local minimum.

This paper also formulates the weakly supervised detector learning as a MIL task. Different from previous works, we introduce a confidence loss term in MIL problem when determining the classifier hyperplane. The key insight is that due to the occlusion, illumination variation, and viewpoint variation, it is suboptimal to treat instances from different bags equally important for detector learning. The introduced confidence loss term measures the reliability of each instance for MIL learning. As a result, the detectors are able to focus on more confident samples and downweights those samples with lower reliability. Furthermore, a cross-validation strategy is introduced to avoid overfitting the initial patterns.
\subsection{Mid-level Image Representation}
A collection of detector responses can be used as mid-level image representation. The paradigm is inspired by object bank \cite{li2010object}, a pioneering work of using detector responses for image representation. The object bank represents an image as a scale-invariant response map of a large number of pre-trained generic object detectors. Following that, most technologies employ detection scores as image representation, and improve the performance by incorporating part responses \cite{shih2015learning}, \cite{singh2012unsupervised} or via multiple scale pooling \cite{parizi2014automatic}, \cite{sun2013learning}.

Over the past years, CNN has become a powerful tool for image representation. Due to the domain mismatch between ImageNet (the source dataset where CNN is trained from) and the target dataset, previous works attempt to enhance CNN representation by transferring learning \cite{oquab2014learning}, \cite{shu2015weakly}, or network fine tuning \cite{girshick2014rich}, \cite{zhang2014fused}. However, these methods need substantial object\,/\,part annotations of the target dataset, which is tedious and impractical in real applications. Zhang \emph{et. al} \cite{zhang2016picking} propose an alternative method to fine tune the network via saliency-based sampling, which is free of the object annotations. Nevertheless, such method is only limited to datasets with relatively simple backgrounds (such as fine grained dataset \cite{WahCUB_200_2011}). It may obtain limited performance improvement on datasets with complex scenes such as Pascal VOC \cite{everingham2010pascal} datasets.

Our approach follows the pipeline of using detector responses as feature representation. Different from previous works which learn a large number of detectors for classification \cite{juneja2013blocks}, \cite{li2015mid}, \cite{parizi2014automatic} or focus on learning a single detector for localization \cite{cinbis2014multi}, \cite{mairal2008discriminative}, \cite{song2014weakly}, \cite{wang2014weakly}, this paper integrates classification and localization into a whole framework, \emph{i.e.}, we not only solve the problem of whether an object is present in an image, but also focus on where the object (if exists) is. We find that it is possible to use only a few detectors for both classification and localization if each detector is distinctive enough. Such an integrated framework is beneficial to close the gap between these two tasks.

Our feature representation is also related to dictionary learning methods \cite{elad2006image}, \cite{mairal2008discriminative}, \cite{zuo2014learning}, where patches are encoded as a sparse linear combination of dictionary elements, optimized for image reconstruction \cite{elad2006image} or recognition \cite{mairal2008discriminative}, \cite{zuo2014learning}. Compared with these approaches, this paper uses detectors as dictionary elements (basis), and chooses detection responses as the combinational coefficients.

\section{Learning Part Detectors}
\begin{figure*}[t]
  \centering
  \includegraphics[width=0.95\textwidth]{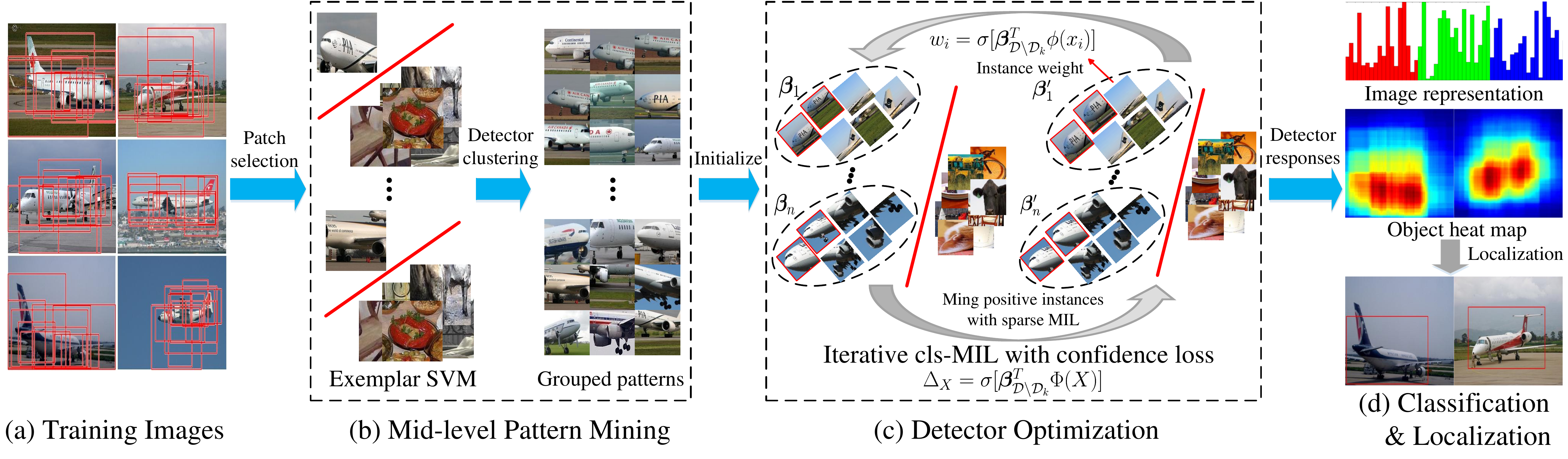}\\
  \vspace{-0.2cm}
  \caption{The framework of the proposed approach. Given a set of training images, we first learn a set of exemplar-SVM detectors from the selected discriminative patches, followed by detector clustering to discover patterns which are consistent and discriminative. The mined patterns are seeded for detector learning, which we formulate as a cls-MIL task. The detector responses are applied for both image classification and object localization. }\label{flowchart}
  \vspace{-0.2cm}
\end{figure*}

In this section, we target at learning a collection of discriminative part detectors automatically for image representation. Our detector learning system consists of two modules: mid-level pattern mining and detector optimization. The pattern mining module first selects patches which are representative and discriminative, then a series of exemplar-SVM \cite{malisiewicz2011ensemble} detectors are trained from each selected patch. This is followed by a spectral clustering procedure which groups exemplar-SVM detectors for pattern mining. Furthermore, an entropy coverage criterion is proposed to measure the generalization ability of each cluster. The detector optimization module formulates the weakly supervised detector training as a confidence loss sparse MIL (cls-MIL) task, which considers the reliability of each positive sample via alternating between mining new positive samples and retraining the part model. The whole framework of the proposed approach is illustrated in Fig. \ref{flowchart}. In the following, we present the detailed design for each module.


\subsection{Pattern Mining with Spectral Clustered Detectors}
Discovering groups of mid-level patterns that are discriminative and representative is crucial for detector learning. To solve this issue, we first introduce a sampling strategy which aims at selecting the discriminative patches, and propose a detector-based spectral clustering approach to mine consistent patterns. Furthermore, we present an entropy coverage criterion to measure the discriminativeness of each cluster, which enables us to greedily select detectors for image representation. These steps are described as follows:

\subsubsection{Discriminative patch selection} It is a challenging task to find discriminative patches without object\,/\,part annotations. To address this issue, a sampling strategy is introduced to select the discriminative and representative patches. Specifically, given an image $I$, we first generate $M$ region proposals $X=\{x_1,...,x_M\}$ with edge boxes \cite{zitnick2014edge}, which probably includes the object of interest with a high recall. Denote the features extracted from a CNN (after ReLU layer) as $\{\phi(I, x_1),...,\phi(I, x_M)\}$, and the final representation of image $I$ is obtained by sum pooling the features over $M$ regions: \emph{i.e.}, $\phi(I)=\frac{1}{M}\sum_{m=1}^M\phi(I, x_m)$. Finally, a one-vs-all SVM classifier is trained based on the sum pooled features $\phi(I)$. Benefit from the non-negativity of CNN features and the additivity of linear classifier, we select the patches which contribute significantly to the classification score. Specifically, given one category $c$ and its classification model $\bm{\beta}_c$, the discriminative patch set $X_D$ of an image $I$ is denoted as:
\begin{equation}\label{dis_patch}
X_D=\{x_i\;|\;\bm{\beta}_c^T\phi(I, x_i)>\tau\},
\end{equation}
where $\tau$ denotes the threshold (set as 1) which enforces selecting the discriminative patches for classification.

\begin{figure}[t]
  \centering
  \includegraphics[width=0.4\textwidth]{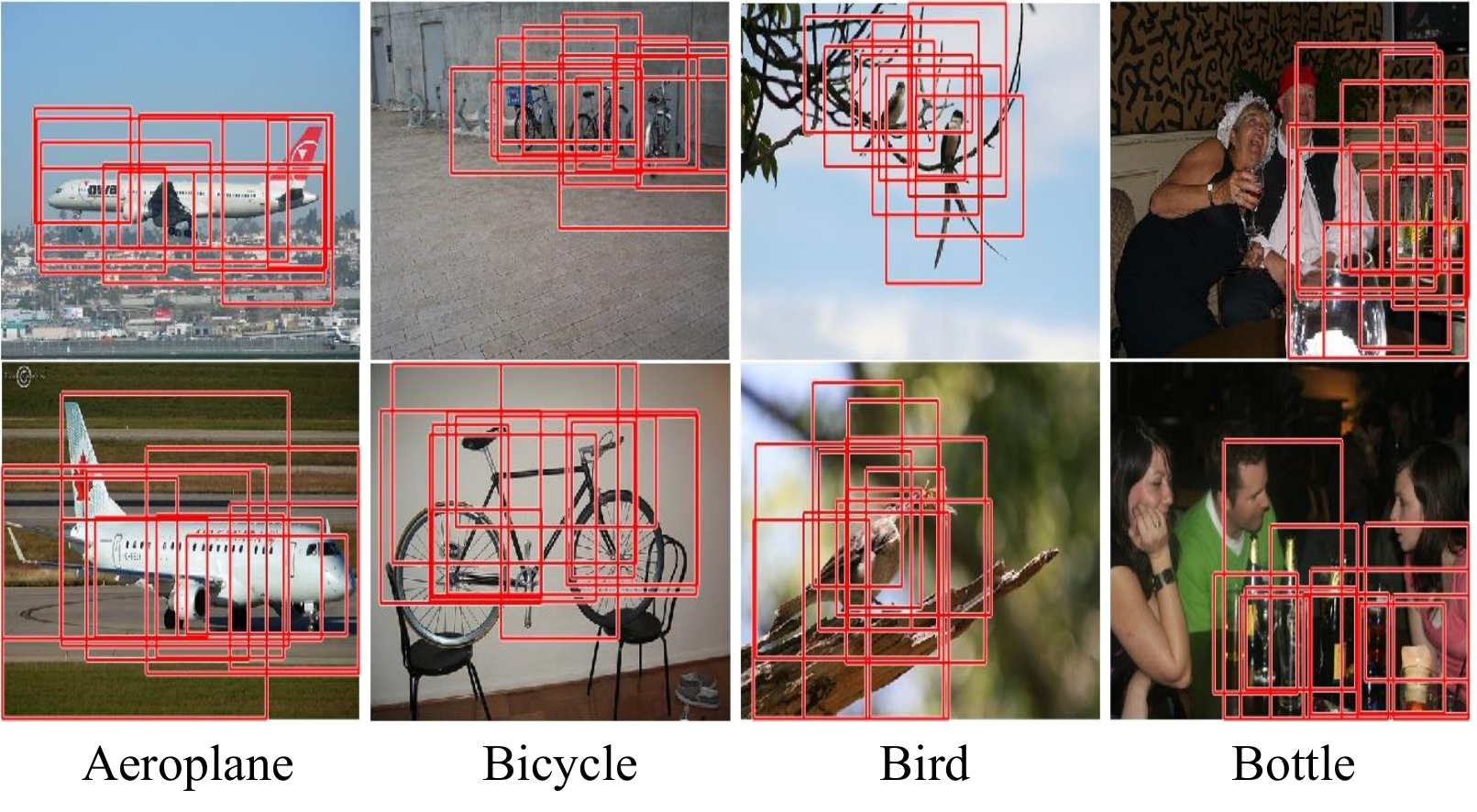} \\
  \vspace{-0.2cm}
  \caption{Examples of the selected discriminative patches (shown in red bounding boxes) on Pascal VOC 2007 \cite{everingham2010pascal}.} \label{pattern_mining}
\vspace{-0.2cm}
\end{figure}

In order to avoid the classifier overfitting the training set $\mathcal{D}$, we equally divide $\mathcal{D}$ into $K$ disjoint and complementary subsets $\mathcal{D}=\{\mathcal{D}_1, ... , \mathcal{D}_K\}$. The classifier is trained on $K\!-\!1$ subsets and validated on the rest one. For generalization, only correctly classified images are retained for discriminative patch selection. Fig. \ref{pattern_mining} illustrates some discriminative patches selected on Pascal VOC 2007 dataset. It can be seen that the selected patches probably locate around the object of interest, and skip other irrelevant backgrounds.

\subsubsection{Detector-based clustering}
The patch selection process usually generates tens of thousands of patterns per category, and most of them are highly correlated, \emph{e.g.}, there exists some patches describing the head of dogs, and some others describing the legs of dogs. It is necessary to cluster these patterns into smaller and representative groups for detector initialization. To this end, an alternative method is to employ some form of unsupervised clustering such as $k$-means \cite{parizi2014automatic}, \cite{singh2012unsupervised}, \cite{sun2013learning}. However, $k$-means behaves poorly in high dimensional space since distance metric becomes less meaningful, and often produces clustered instances which are in no way visually similar. Instead of clustering the original patches, this paper proposes a detector-based spectral clustering strategy, which discovers similar patterns via the grouped detectors.

\begin{figure*}[t]
  \centering
  \includegraphics[width=0.9\textwidth,height=8cm]{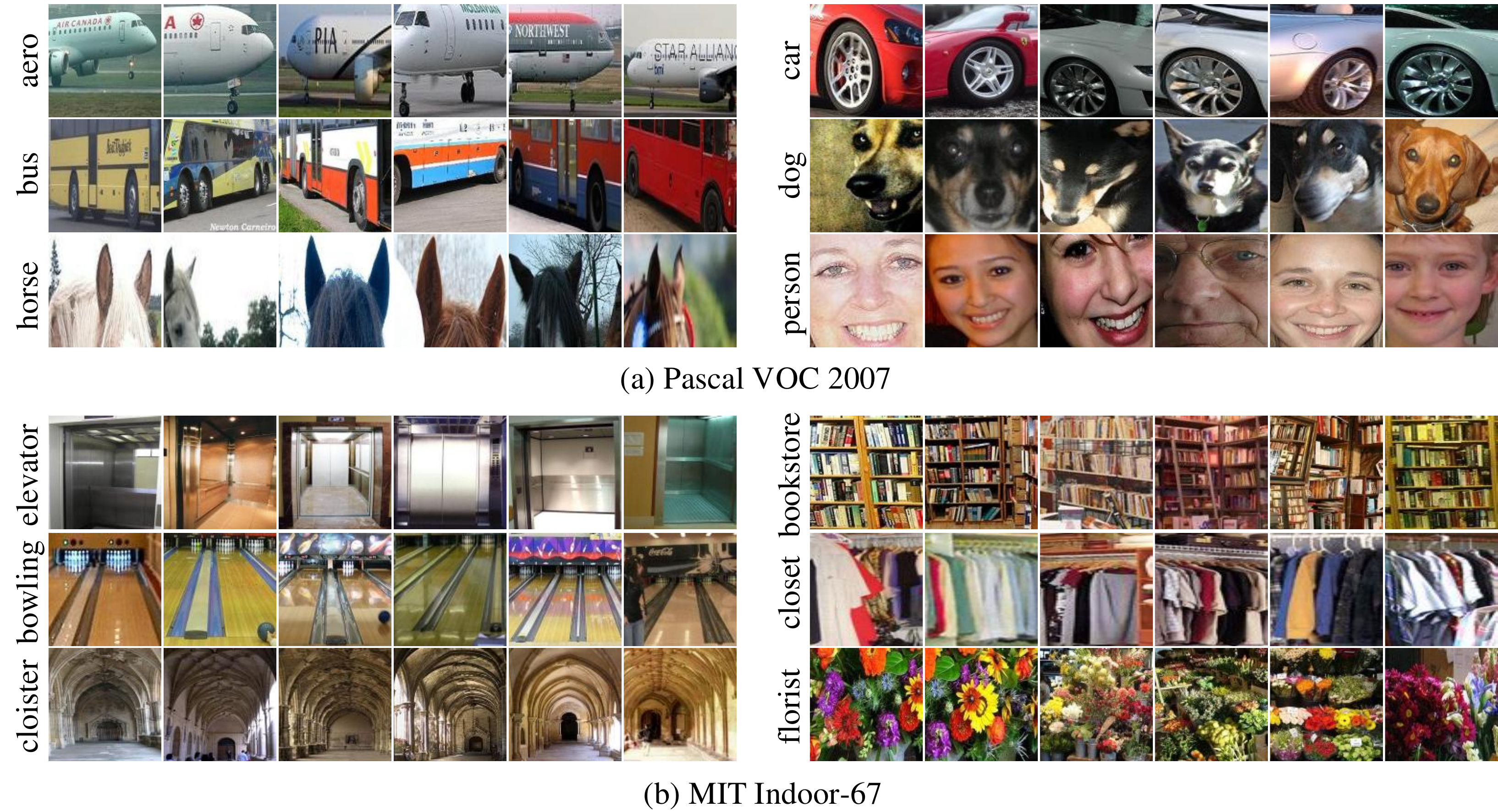}\\
  \vspace{-0.2cm}
  \caption{Examples of the discovered mid-level patterns with clustered detectors on (a) Pascal VOC 2007 \cite{everingham2010pascal} and (b) MIT Indoor-67 \cite{quattoni2009recognizing}. These patterns are obtained by the top responses of each clustered detectors. } \label{cluster_patterns}
\vspace{-0.2cm}
\end{figure*}

Inspired from exemplar-SVMs \cite{malisiewicz2011ensemble}, we start learning detectors from only one instance, which avoids worrying about exemplar misalignment. The negative samples are defined as patches which do not contain the object of interest, \emph{i.e.}, all patches sampled from images with different labels. Since the negative samples are too large, standard hard mining method \cite{felzenszwalb2008dpm} is quite expensive. We use instead Linear Discriminant Analysis (LDA) \cite{hariharan2012discriminative} to train a detector, which is an accelerated version of the exemplar-SVMs. Specifically, the detector template $\bm{d}$ is learned simply by $\bm{d}=\sum^{-1}(\bm{\overline{x}}_p-\bm{\mu}_0)$, where $\bm{\overline{x}}_p$ is the mean features of the positive examples, $\bm{\mu}_0$ denotes the mean of the features in the whole dataset, and $\sum$ is the corresponding covariance matrix. Since each exemplar-SVM detector is supposed to fire only on visually similar examples, we cannot expect it to generalize too much. To solve this issue, we follow an iterative procedure \cite{juneja2013blocks} which adds new positive samples each round to enhance the exemplar detectors. At each round, we run the current detector on all other images with the same label, and retrain it by augmenting the positives with the top scored patches. The idea behind this process is using detection score as a similarity metric, which emphasizes the distinctive details and suppresses those irrelevant ones.

Using exemplar-SVMs, each selected patch is associated with a detector. The key insight of the proposed strategy is that instead of clustering the original patches, we group the corresponding detectors. Specifically, given $n_c$ exemplar detectors $\{\bm{d}_i\}_{i=1}^{n_c}$ trained from one class $c$, we perform spectral clustering on the similarity matrix $S$ generated from the detectors, and obtain $\mathcal{K}$ clusters $\{C_k\}_{k=1}^\mathcal{K}$, where $S(i,j)$ denotes the cosine similarity of $\bm{d}_i$ and $\bm{d}_j$. Thus, detectors sharing similar response distributions are grouped together. Inspired by boosting strategy \cite{viola2004robust}, each cluster acts as an integrated detector to discover similar patterns, \emph{i.e.,} the detection score of a patch $x$ with respect to a cluster $C_k$ is denoted as:
\begin{equation}\label{entropy_score}
   s(C_k|x)=\sum_{\bm{d}_k \subseteq C_k} \bm{d}_k \phi(x).
\end{equation}
As an illustration, Fig. \ref{cluster_patterns} shows some examples of the discovered patterns using the clustered detectors. It can be shown that although a single detector is weak, a collection of such detectors offer satisfactory localization capacity. Another advantage of the detector-based pattern mining method is that we can select the most discriminative and representative patterns according to the top responses of the grouped detectors.
\subsubsection{Entropy coverage}
The detector-based clustering generates a series of clusters with varying discriminative capacities. The notation of discriminative clusters is that the detectors within a cluster should be trained from as many images as possible. Such clusters include detectors corresponding to repeated patterns among varying images. We propose an entropy coverage criterion to measure the discriminativeness of each cluster. Given $N$ images $\{I_i\}_{i=1}^N$ belonging to the same class and the corresponding $\mathcal{K}$ clustered detectors $\{C_k\}_{k=1}^\mathcal{K}$, the entropy coverage of cluster $C_k$ is defined as:

\begin{equation}\label{entropy_coverage}
  \mathcal{H}(C_k)=-\sum_{i=1}^N p(I_i|C_k)\log_2{p(I_i|C_k)},
\end{equation}
where $p(I_i|C_k)$ denotes the probability of detectors coming from image $I_i$. The subitem of $\mathcal{H}(C_k)$ is a standard entropy function, which enjoys the following property:

\textbf{Corollary 1.} Denote the entropy function as $H(p)=-p\log_2p$, then for $\forall \, \, 0<p<1, \, 0< \Delta p < p, H(p)<H(p-\Delta p)+H(\Delta p) \leq 2H(\frac{p}{2})$.

\textit{Proof.} For the left side, we have:
\begin{equation}\label{entropy_demo}
 \begin{split}
  H(p)&=-p\log_2(p) \\
  &=-(p-\Delta p)\log_2(p)-\Delta p \log_2(p) \\
  &<-(p-\Delta p)\log_2(p-\Delta p)-\Delta p \log_2(\Delta p) \\
  &=H(p-\Delta p)+H(\Delta p). \\
  \end{split}
\end{equation}
The right side is obvious according to the maximum property of entropy, \emph{i.e.}, the entropy reaches its maximum when events are equiprobable.

According to \textbf{Corollary 1}, $\mathcal{H}(C_k)$ is large if the clustered detectors within $C_k$ are trained from diverse images, and reaches its maximum when the detectors are trained from patterns with equal distribution. The larger $\mathcal{H}(C_k)$ is, the more frequent patterns the detectors in $C_k$ could find. Such an entropy coverage criterion enables us to greedily select clusters for detector initialization, while not worrying about choosing appropriate number of clusters. In the experimental section, we would find that the optimal number of clusters is determined by the classification performance.
\subsection{Detector Optimization with cls-MIL}
Although the grouped detectors offer a relatively robust localization capacity, it is far from enough. These detectors are trained from a subset of discriminative patches, and are only powerful to discover patches which are also significant in discriminativeness. While we cannot ensure that they respond consistently among all the images of that class, especially those not correctly classified ones in cross-validation. Based on these observations, we formulate the weakly supervised detector learning as a confidence loss sparse MIL (cls-MIL) task, which considers the diversity of the positive samples, while avoid drifting away the well localized ones by assigning a confidence value to each mined positive sample.
\subsubsection{Motivation}
To use MIL for detector learning, each image is considered as a bag, and the patches within it as instances. Given a set of training images, we treat images of one particular category as positive bags, and the rest images as negative bags. Intuitively, for each image, if it is labeled as positive, then at least one patch within it should be treated as a positive instance, when it is labeled as negative, then all patches within it should be treated as negative instances. Standard MIL is based on alternatively selecting the highest detection per bag as the positive instance and refining the detection model. However, it suffers from several issues. First, the detectors would latch on to the initial patches they are trained from and prefer them at each round of instance selection when training and selecting are performed on the same dataset. Second, standard MIL often mines a single instance per positive bag and treats each mined instance equally important, which is often not the case. Due to the occlusion, illumination variation, and viewpoint variation, the same part from different images suffers from varying confidence of positiveness. Based on these observations, a multi-fold cross-validation \cite{cinbis2014multi} is introduced to avoid overfitting the initial training samples, and a confidence loss sparse MIL (cls-MIL) technology is proposed to tackle the dataset bias. In the following we define the problem in a formal way.

\subsubsection{Problem formulation}
Let $\mathcal{X}$ be the set of bags used for training, which consists of a set of positive bags $\mathcal{X}_p$ and negative bags $\mathcal{X}_n$, \emph{i.e.}, $\mathcal{X}=\mathcal{X}_p \cup \mathcal{X}_n$. Denote $X$ be a bag of images, and $\tilde{\mathcal{X}}_p=\{x|x \in X \subseteq \mathcal{X}_p\}$ and $\tilde{\mathcal{X}}_n=\{x|x \in X \subseteq \mathcal{X}_n\}$ be the set of instances from positive bags and negative bags, respectively. For any instance $x \in X$ from a bag $X \subseteq \mathcal{X}$, let $\phi(x)$ be the feature vector representation of $x$ (for brevity, we include the bias term into feature representation). The cls-MIL problem can be formulated as solving the following objective:
\begin{equation}\label{mil_loss}
\begin{split}
  \min & \quad \frac{1}{2}||\bm{\beta}||^2+C\sum_{X \subseteq \mathcal{X}_p}\Delta_X \xi_X+C\sum_{x \in \tilde{\mathcal{X}}_n} \xi_x \\
 \textit{s.t.} & \quad \bm{\beta}^T\Phi(X) \geq 1-\xi_X, \ \ \forall X \subseteq \mathcal{X}_p,\\
  & \quad \bm{\beta}^T\phi(x) \leq -1+\xi_x, \ \ \forall x \in \tilde{\mathcal{X}}_n,\\
  & \quad \xi_X \geq 0, \quad \xi_x \geq 0, \ \ \forall X \subseteq \mathcal{X}_p, \forall x \in \tilde{\mathcal{X}}_n,
  \end{split}
\end{equation}
where $\Phi(X)$ is the feature representation of bag $X$, $\Delta_X$ is the latent variable which measures the positiveness of a bag $X \subseteq \mathcal{X}_p$, and $C$ is the control parameter of the loss term.

One remained issue is how to determine the representation $\Phi(X)$. We would prefer that a positive bag be represented as much as possible by the true positives within it. However, even the state-of-the-art region proposal algorithms \cite{zitnick2014edge} could only generate patches containing the object of interest with a high recall, not to mention the difficulty of determining the positive samples under weakly supervised paradigm. To tackle this issue, we introduce a pooling strategy for $\Phi(X)$ representation to improve the robustness. Note that among all the given region proposals, only a few instances are the patterns we expect to find (which is sparse). Based on these observations, each bag is represented as the weighted sum of its mined member instances: $\Phi(X)=\frac{\sum_{m \in s(X)} w_m\phi(x_m)}{\sum_{m \in s(X)} w_m}$, where $w_m$ is a weight assigned to each instance, and $s(X)$ is an indicator which denotes the patterns selected as the positive ``witness'' in a positive bag $X$. In practice, only a few instances per positive bag are selected (we set the number of $s(X)$ as 10), while all the negative instances are taken into consideration.


\begin{algorithm}[t]
\caption{Weakly Supervised Detector Learning}\label{algorithm1}
\begin{algorithmic}
\REQUIRE  Positive bags $\mathcal{X}_p$, negative bags $\mathcal{X}_n$, the number of spectral clusters $\mathcal{K}$, and the number of iterations T; \\
\textbf{Mid-level Pattern Mining:} For instances in the positive bags $\mathcal{X}_p$, mining patterns for detector initialization.

a). Select discriminative patches $\{x_i\}_{i=1}^m$ with Eq. (\ref{dis_patch}) via cross-validation.\\
b). For each selected patch $x_p \in \{x_i\}_{i=1}^m$, learn exemplar-SVM detector $\bm{d}=\sum^{-1}(\bm{\overline{x}}_p-\bm{\mu}_0)$.\\
c). Spectral clustering of detectors $\{\bm{d}_i\}_{i=1}^m$ into $\mathcal{K}$ clusters $\{C_k\}_{k=1}^\mathcal{K}$. \\
d). For each cluster, pattern mining on $\mathcal{X}_p$ according to scores $s(C_k|x)\!=\!\sum_{\bm{d}_k \subseteq C_k}\! \bm{d}_k \phi(x)$. \\
\textbf{Detector Optimization:} For each cluster, given initial patterns discovered by the clustered detectors, solving cls-MIL in Eq. (\ref{mil_loss}) via iteratively updating and optimizing. \\
\textbf{For} iteration t=1 to T

a). Updating: Updating the latent variables via cross-validation. The latent variables in $\mathcal{D}_k$ are determined by detectors $\bm{\beta}_{\mathcal{D}\setminus\mathcal{D}_k}$ trained on  $\{\mathcal{D}\!\setminus\!\mathcal{D}_k\}$, \emph{i.e.}, updating instance weights $w_m$ of $\Phi(X)$ by: $w_m= \sigma[\bm{\beta}^T_{\mathcal{D}\setminus\mathcal{D}_k}\phi(x_m)]$, and the confidence loss term $\Delta_X=\sigma[\bm{\beta}^T_{\mathcal{D}\setminus\mathcal{D}_k}\Phi(X)]$. \\
b). Optimizing: solving Eq. (\ref{mil_loss}) via hard negative mining on negative bags $\mathcal{X}_p$ with the updated latent variables $\Phi(X)$ and $\Delta_X$. \\
\textbf{end}
\ENSURE Detector set $\{\bm{\beta}_k\}_{k=1}^{\mathcal{K}}$.
\end{algorithmic}
\end{algorithm}

\subsubsection{Optimization}
The cls-MIL leads to a non-convex optimization problem due to the introduction of implicit feature representation $\Phi(X)$ for the positive bags and the latent confidence variables $\Delta_X$. However, this problem is semi-convex since optimization problem becomes convex once these latent variables are fixed. In the following, we solve Eq. (\ref{mil_loss}) via an iterative procedure which alternates between fixing the latent variables and optimizing the detectors. In order to avoid focusing on the initial positive samples, the optimization procedure is processed via cross-validation. Specifically, the training set $\mathcal{D}$ is equally divided into $K$ disjoint and complementary subsets $\{\mathcal{D}_1, ... , \mathcal{D}_K\}$. Starting from the patterns discovered by the clustered exemplar-SVM detectors, the detector $\bm{\beta}$ is optimized via iteratively \textbf{Updating} the latent variables and \textbf{Optimizing} Eq. (\ref{mil_loss}). In the \textbf{Updating} step, the latent variables in $\mathcal{D}_k$ are determined by detectors $\bm{\beta}_{\mathcal{D}\setminus\mathcal{D}_k}$ trained on  $\{\mathcal{D}\!\setminus\!\mathcal{D}_k\}$, \emph{i.e.}, each instance weight $w_m$ of $\Phi(X)$ is updated by: $w_m= \sigma[\bm{\beta}^T_{\mathcal{D}\setminus\mathcal{D}_k}\phi(x_m)]$, and the confidence loss term $\Delta_X=\sigma[\bm{\beta}^T_{\mathcal{D}\setminus\mathcal{D}_k}\Phi(X)]$, where $\sigma$ is a sigmoid function which maps the value into the range of $(0,1)$. In the \textbf{Optimizing} step, the detector is optimized according to the updated latent variables via hard negative mining \cite{felzenszwalb2008dpm}.

\textbf{Corollary 2.} The solution $\bm{\beta}$ of Eq. (\ref{mil_loss}) is a linear combination of the positive instances $\phi(X)$ and the negative instances $\phi(x)$, \emph{i.e.,} $\bm{\beta}=\sum_{X \subseteq \mathcal{X}_p} \alpha_X\phi(X)+\sum_{x \in \tilde{\mathcal{X}}_n} \alpha_x\phi(x)$, where the coefficients $\alpha_X$ and $\alpha_x$ are bounded by: $0 \leq \alpha_X \leq C \Delta_X$, $0 \leq \alpha_x \leq C$, respectively.

\emph{Proof.} The constrained minimization problem in Eq. (\ref{mil_loss}) can be solved with a classical Lagrangian method. The Lagrangian operator can be represented as:
\begin{equation}\label{lagrangian}
\begin{aligned}
\mathscr{L}&=\frac{1}{2}||\bm{\beta}||^2+C\sum_{X \subseteq \mathcal{X}_p}\Delta_X \xi_X+C\sum_{x \in \tilde{\mathcal{X}}_n}\xi_x\\
&+\alpha_x(\bm{\beta}^T\phi(x)+1-\xi_x)-
\sum_{x \in \tilde{\mathcal{X}}_n}\gamma_x\xi_x \\
&-\alpha_X(\bm{\beta}^T\Phi(X)-1+\xi_X)
-\sum_{X\subseteq \mathcal{X}_p}\gamma_X\xi_X,
\end{aligned}
\end{equation}
where $\alpha_X$, $\alpha_x$, $\gamma_X$, and $\gamma_x$ denote Lagrange multipliers. The minimization of Lagrangian operator in Eq. (\ref{lagrangian}) with respect to $\bm{\beta}, \xi_X, \xi_x$ is obtained:
\begin{equation}
\begin{cases} \vspace{0.1cm}
\frac{\partial \mathscr{L}}{\partial \bm{\beta}}\!=\!0\Rightarrow \bm{\beta}\!=\!\sum_{X \subseteq \mathcal{X}_p} \alpha_X \phi(X)\!-\!\sum_{x \in \tilde{\mathcal{X}}_n}\alpha_x\phi(x), \\ \vspace{0.15cm}
\frac{\partial \mathscr{L}}{\partial \xi_X}\!=\!0 \Rightarrow \gamma_X\!=\!\Delta_XC\!-\!\alpha_X, \\ \vspace{0.15cm}
\frac{\partial \mathscr{L}}{\partial \xi_x}\!=\!0 \Rightarrow \gamma_x\!=\!C\!-\!\alpha_x. \\
\end{cases}
\end{equation}
Due to the nonnegativity of $\gamma_X$ and $\gamma_x$, we have $0 \leq \alpha_X \leq C \Delta_X$ and $0 \leq \alpha_x \leq C$. Given a test example $\tilde{x}$, the detection score can be represented as:
\begin{equation}\label{detection_score}
 f(\tilde{x})\!=\!\left(\sum_{X \subseteq \mathcal{X}_p} \alpha_X \phi(X)\!-\!\sum_{x \in \tilde{\mathcal{X}}_n}\alpha_x\phi(x)\right)\phi(\tilde{x}), \\
\end{equation}
It can be seen that the final detection score $f(\tilde{x})$ is a weighted combination of the inner product between training features $\phi(X)$, $\phi(x)$ and test feature $\phi(\tilde{x})$, and is only determined by samples with nonzero coefficients $\alpha_i \, (i=X,x)$. These $\alpha_i$s are called support vectors, since they are the only training samples necessary to define the separating hyperplane. Note that for positive samples, the coefficient $\alpha_X$ is bounded by $C\Delta_X$, with KKT conditions, it is also possible to see when an example is a support vector, this happens only if the example is on the margin, or it does not respect the separation conditions in Eq. (\ref{mil_loss}). According to \cite{collobert2004large}, the coefficient $\alpha_X$ for positive samples in different locations is defined as:
\begin{equation}\label{alpha_value}
\begin{cases}
\,\, \alpha_X=0, \quad \bm{\beta}^T\phi(X)>1, \\
\,\, \alpha_X=C\Delta_X, \quad \bm{\beta}^T\phi(X)<1, \\
\,\, 0 < \alpha_X < C\Delta_X, \quad \bm{\beta}^T\phi(X)=1. \\
\end{cases}
\end{equation}

For positive bags which do not respect the classification hyperplane, the corresponding coefficient $\alpha_X$ is bounded by $C\Delta_X$, which takes the reliability of $X$ into consideration. The regularized term $\Delta_X$ helps to boost the detection performance. If a positive bag $X$ is not reliable at previous round, its contribution to the classification hyperplane at current round would be lowered. As a result, MIL introduces diverse samples for detector learning, while the confidence loss term encourages the detector focusing on positive instances which are good enough and downweighting those instances with lower reliability. The whole procedure of the proposed weakly supervised detector learning algorithm is summarized in Algorithm \ref{algorithm1}.
\begin{figure}[t]
  \centering
  \includegraphics[width=0.45\textwidth]{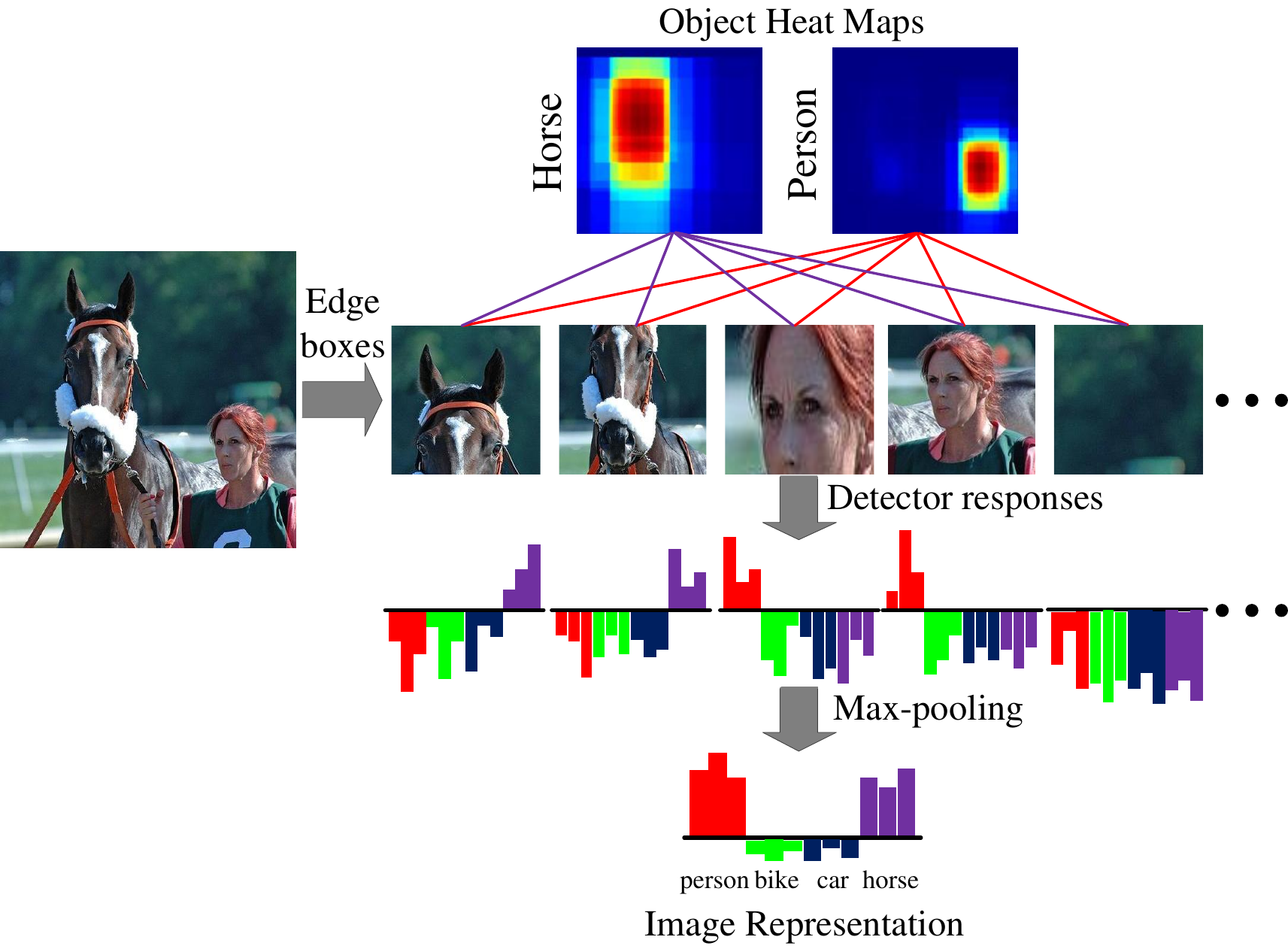}\\
  \vspace{-0.1cm}
  \caption{An illustration of how to compute image representation and object heat maps according to the detector responses.}\label{cls_loc}
  \vspace{-0.2cm}
\end{figure}
\section{Applications: Image Classification and Object Localization}
The learned detectors are discriminative for the corresponding category, and an ensemble of the detectors across different categories offers an effective mid-level image representation. In this section, we apply such mid-level representation for image classification and object localization.
\subsection{Image Classification}
Unsupervised clustering methods have been used for feature representation \cite{perronnin2010improving}, \cite{wang2010locality}. Since our learned detectors can be considered as the true visual patterns corresponding to a certain category (as opposed to the clustered ambiguous visual letters in \cite{perronnin2010improving}, \cite{wang2010locality}), it makes sense to apply such detectors for image coding. Denote all the learned detectors across different categories as $\Gamma=\{\bm{\beta}_i\}_{i=1}^K$, where $K$ is the total number of detectors. Our mid-level feature representation is based on the maximal responses of a collection of detectors. Specifically, given an image $I$ and the corresponding region proposals $X$, the feature representation is denoted as:
$f(I, \Gamma)= [\bm{\beta}_1^T\phi(I, z_1),\, ... \, , \, \bm{\beta}_K^T\phi(I, z_{K})]$, where $z_k$ is a latent variable indicating the region with maximum response corresponding to detector $\bm{\beta}_k$, \emph{i.e.}, $z_k=\text{argmax}_{z \in X} \bm{\beta}_k^T\phi(I, z)$. An illustration of image representation is shown in Fig. \ref{cls_loc}.
\begin{figure*}[t]
  \centering
  \includegraphics[width=0.95\textwidth]{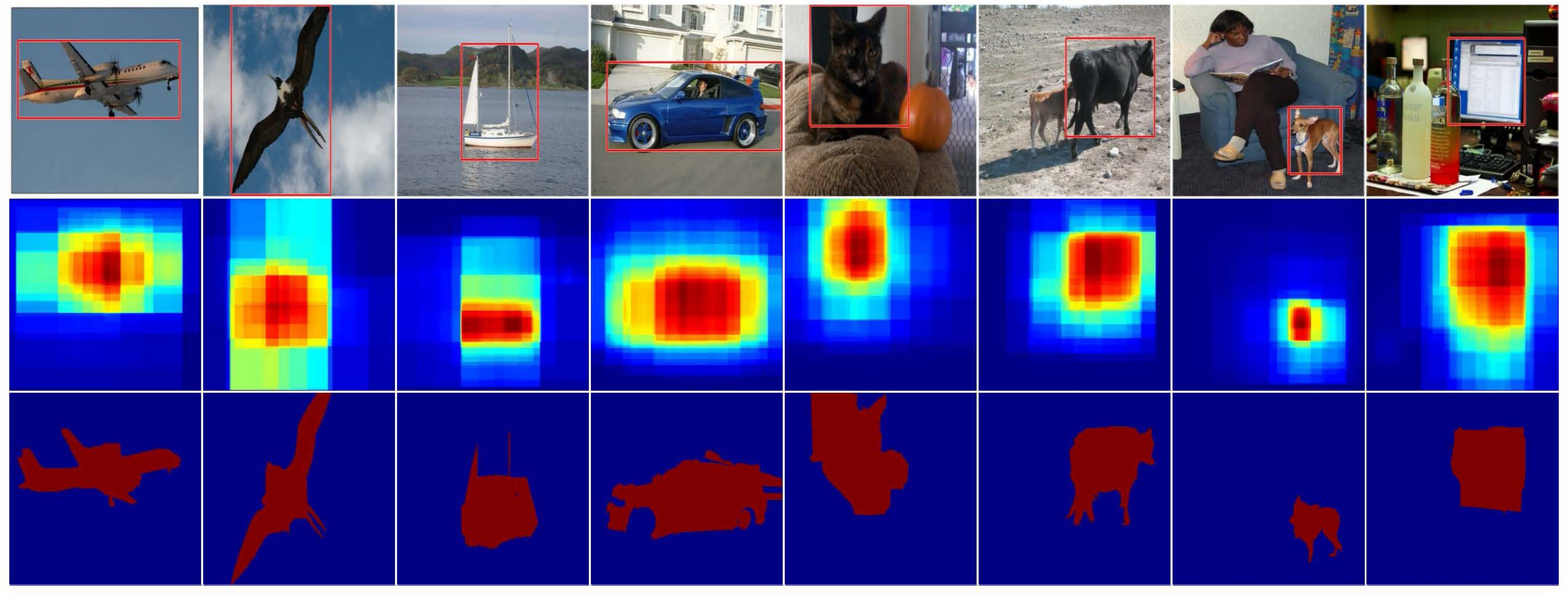} \\
  \vspace{-0.2cm}
  \caption{Examples of localization process on Pascal VOC 2007 \emph{trainval} split. We generate the object heat map and perform grabcut \cite{rother2004grabcut} to obtain segmentation mask of the object. Then a tight object bounding box (shown in red) is obtained via enclosing the segmentation mask.} \label{detection_illus}
\vspace{-0.2cm}
\end{figure*}

Given the image representation, a conventional SVM classifier is performed to produce the final classification results. Note that the complexity of the feature coding using detector responses is very low, which includes no more than a dot product operation once the features (\emph{e.g.}, CNN) are extracted. On the other hand, we greedily select detector responses based on the entropy coverage criterion, and find that the performance saturates as the first few detectors are added in, which decreases the feature dimension by one order. In the experimental section, we will demonstrate the effectiveness of the proposed feature coding approach.

\subsection{Object Localization}
The learned part detectors are discriminative for the corresponding category, and a collection of them offers rough position of the object of interest. In this section, we present a simple object localization technology based on the learned part detectors. The basic idea is to accumulate the part responses into a whole object heat map, which indicates the potential position of an object. Specifically, starting from a collection of part detectors $\{\bm{\beta}_k\}_{k=1}^\mathcal{K}$ corresponding to a category, we first define a part map $\mathcal{O}_k$ based on detector $\bm{\beta}_k$, the confidence of a pixel $p$ which is contained in an object part is denoted as:
\begin{equation}\label{dis_map}
\mathcal{O}_k(p)=\frac{\sum_{x_m \subset \Omega_p} \sigma[\bm{\beta}_k^T\phi(I, x_m)]}{Z},
\end{equation}
where $\Omega_p$ denotes the patch set that includes pixel $p$, $\sigma$ is a sigmoid function, and $Z$ is a normalization constant such that $\max \mathcal{O}_k(p)\!=\!1$. Finally, the object map is a weighted linear combination of the part maps obtained by all part detectors, \emph{i.e.}, $\mathcal{O}(p)=\sum_{k=1}^\mathcal{K}\frac{w_k\mathcal{O}_k(p)}{\sum_k w_k}$, where $w_k$ is a weight factor which denotes the reliability of each detector, and is given by $w_k=\max_{m \in X} \sigma[\bm{\beta}_k^T\phi(I, x_m)]$. Fig. \ref{cls_loc} illustrates examples of how to compute the object heat maps.

The object heat map indicates the most discriminative details of an object, and usually focuses on object parts (\emph{e.g.}, the head of dogs), instead of the whole object. Inspired from \cite{parkhi2011truth} which casts localization as a segmentation task, we perform grabcut \cite{rother2004grabcut} on the object heat map to generate the segmentation mask. The goal is to propagate the discriminative part details to the whole object with color continuity cues. To this end, the foreground and background are set to be gaussian mixture models. The foreground model is estimated from heat map values higher than $0.8$, and the background model is estimated from values lower than $0.2$. Finally, we take the bounding box that covers the largest connected component in the generated segmentation mask as localization result. Some example localization processes are shown in Fig. \ref{detection_illus}. In the experimental section, we will show that as a byproduct of the learned discriminative detectors, such localization technique achieves satisfactory localization performance.
\section{Experiments}
In this section, we present an evaluation of the proposed weakly supervised image classification and object localization framework. We also perform ablation study to understand how various design choices impact the recognition performance.
\subsection{Datasets and evaluation metrics}
We evaluate the proposed approach on three publicly available benchmarks, two for generic object recognition and one for scene recognition. The details of the datasets are briefly summarized as follows:

\textbf{Pascal VOC 2007:} The Pascal VOC 2007 dataset \cite{everingham2010pascal} is a widely used benchmark for multi-label image classification and object localization.  The benchmark contains a total of 9,963 images spanning 20 generic object classes, of which 5,011 images are used for \emph{trainval} and the rest 4,952 images for \emph{test}. For image classification, we choose \emph{trainval} split as training set and \emph{test} split as test set, and the evaluation metrics is mean Average Precision (mAP), which is complying with the Pascal VOC challenge protocols.

\textbf{Pascal VOC 2012:} The Pascal VOC 2012 dataset \cite{everingham2015pascal} is an extended version of the Pascal VOC 2007, which contains a total of 22,531 images, including 11,540 images for \emph{trainval} and 10,991 images for \emph{test}. Since ground truth labels are not available for \emph{test} split, we use the online evaluation server to evaluate recognition performance of the proposed algorithm.

\textbf{MIT Indoor-67:} The MIT Indoor-67 \cite{quattoni2009recognizing} dataset consists of 15,620 images belonging to 67 categories of indoor scenes. It is challenging because of the large ambiguities between categories. We follow the standard \emph{train}/\emph{test} split as in \cite{quattoni2009recognizing}, \emph{i.e.}, approximately 80 images per class for \emph{train} and 20 images per class for \emph{test}. The evaluation metric for MIT Indoor dataset is the mean classification accuracy.

In addition to classification, we also evaluate the localization performance of the proposed approach. We follow previous methods on object localization \cite{Bilen_2016_CVPR}, \cite{cinbis2014multi}, and evaluate the performance on Pascal VOC \emph{trainval} set with CorLoc criterion  \cite{deselaers2012weakly}. CorLoc measures the percentage of images with correct localization, \emph{i.e.}, a window is considered to be correct if it has an Intersection-over-Union (IoU) ratio of at least $50\%$ with one of the ground truth instances.
\begin{table}[t]
\centering
\caption{Recognition performance on VOC 2007 with different number of region proposals. Results are based on model CaffeNet.} \label{no_regions}
\vspace{-0.1cm}
\begin{tabular}{l|c|c|c|c}
\hline
NO. of proposals & 300    & 500    & 1000   &2000   \\ \hline
mAP              & 82.4\% & 83.2\% & 83.4\% & 83.7\% \\ \hline
\end{tabular}
\vspace{-0.1cm}
\end{table}
\subsection{Implementation Details}
$\bullet$ \textbf{Models and features.} We choose two widely used CNN models for feature extraction, a typical network CaffeNet \cite{jia2014caffe} and a more accurate but deeper one VGG-VD \cite{Simonyan14c} (the $16$-layer model). We extract features from the fc6 layer (FC-CNN) after the rectified linear unit (ReLU), which is a $4096$-d nonnegative vector for each region. Edge boxes \cite{zitnick2014edge} are used for generating candidate region proposals. In addition to region proposals, edge boxes also provide an objectness score for each region. For computation efficiency, we disregard regions which occupy less than $1\%$ areas of an image, and retain the top scored $500$ region proposals as candidates.

$\bullet$ \textbf{Parameter settings.} In pattern mining, the number of spectral clustering per category $\mathcal{K}$ is set as $50$, and the top scored $100$ patches per clustered detectors are selected as patterns for detector initialization. In detector optimization, the number of iterations $T$ is set as $3$, as we find that the performance of the detectors do not need more to converge. For all situations where cross-validation is needed, we use typical $5$-fold cross-validation.
\subsection{Ablation Study}
To better understand the relative contribution of each module, we analyze the performance of our approach with different configurations. As the localization can be regarded as a byproduct of the learned detectors, we mainly measure how different designs affect the discriminativeness of the detectors in terms of classification performance.
\begin{table*}[t]
\caption{Recognition average precision ($\%$) on VOC 2007 \emph{test} split. We report performance with two models: CaffeNet \cite{jia2014caffe} and VGG-VD \cite{Simonyan14c}. The method marked with $*$ are those using additional training images.} \label{rec07_results}
\scriptsize
\vspace{-0.1cm}
\setlength\tabcolsep{4.0pt}
\begin{tabular}{l|cccccccccccccccccccc|c}
\hline
method&aero&bike&bird&boat&bottle&bus&car&cat&chair&cow&
table&dog&horse&mbike&persn&plant&sheep&sofa&train&tv&mAP \\ \hline
MR-CaffeNet \cite{jia2014caffe}
&90.4&87.0&87.2&84.1&40.5&76.4&86.9&87.5&60.7&70.5&75.7&82.7&89.4&80.4
&93.9&53.9&76.6&66.6&90.9&71.5&77.6 \\ \hline
MR-VD \cite{Simonyan14c}
&98.3&95.3&96.0&95.0&70.0&90.1&93.8&94.9&73.7&84.6&85.9&94.5&95.4&92.0
&97.5&70.6&90.6&79.7&98.1&86.7&89.1 \\ \thickhline
PRE-1000$^*$\!\cite{oquab2014learning}
&88.5&81.5&87.9&82.0&47.5&75.5&90.1&87.2&61.6&75.7&67.3&85.5&83.5&80.0
&95.6&60.8&76.8&58.0&90.4&77.9&77.7 \\ \hline
HCP Alex$^*$\!\cite{hcp_pami}
&95.4&90.7&92.9&88.9&53.9&81.9&91.8&92.6&60.3&79.3&73.0&90.8&89.2&86.4
&92.5&66.9&86.4&65.6&94.4&80.4&82.7 \\ \hline
HCP VD$^*$\!\cite{hcp_pami}
&98.6&97.1&\textbf{98.0}&95.6&75.3&\textbf{94.7}&95.8&97.3&73.1&90.2&80.0&\textbf{97.3}&96.1
&\textbf{94.9}&96.3&78.3&\textbf{94.7}&76.2&97.9&91.5&90.9 \\ \hline
WSDDN \cite{Bilen_2016_CVPR} &93.3&93.9&91.6&90.8&\textbf{82.5}&91.4&92.9&93.0&\textbf{78.1}&\textbf{90.5}&82.3&95.4&92.7&92.4
&95.1&\textbf{83.4}&90.5&80.1&94.5&89.6&89.7 \\ \hline
\textbf{EPD} CaffeNet
&94.6&92.0&90.4&89.3&56.9&81.9&93.0&90.8&67.9&71.7&77.0&84.9&89.7&86.4
&97.1&71.8&80.7&69.4&93.8&84.3&83.2 \\ \hline
\textbf{EPD} VD
&98.6&97.7&97.2&96.0&78.4&92.0&95.8&96.9&76.5&86.9&82.4&94.1&95.3&93.5
&98.6&79.4&94.5&80.1&98.6&92.2&\textbf{91.3} \\ \hline
\textbf{EPD} VD+\!\cite{Simonyan14c}
&\textbf{99.3}&\textbf{97.8}&97.6&\textbf{96.4}&79.1&92.9&\textbf{95.9}&\textbf{97.3}&78.0&88.5
&\textbf{87.1}&95.4&\textbf{96.1}&94.4&\textbf{98.7}&80.0&94.6&\textbf{82.9}&\textbf{99.0}&\textbf{92.2}
&\textbf{92.2} \\ \hline
\end{tabular}
\vspace{-0.1cm}
\end{table*}

\begin{table*}[t]
\caption{Recognition average precision ($\%$) on VOC 2012 test. The method marked with $*$ are those using additional training images. Available at \href{http://host.robots.ox.ac.uk:8080/anonymous/UKZVBM.html}
{http://host.robots.ox.ac.uk:8080/anonymous/UKZVBM.html} and \href{http://host.robots.ox.ac.uk:8080/anonymous/CD25HO.html}
{http://host.robots.ox.ac.uk:8080/anonymous/CD25HO.html}.} \label{rec12_results}
\scriptsize
\vspace{-0.1cm}
\setlength\tabcolsep{4.1pt}
\begin{tabular}{l|cccccccccccccccccccc|c}
\hline
method&aero&bike&bird&boat&bottle&bus&car&cat&chair&cow&
table&dog&horse&mbike&persn&plant&sheep&sofa&train&tv&mAP \\ \hline
PRE-1000$^*$\!\cite{oquab2014learning}
&93.5&78.4&87.7&80.9&57.3&85.0&81.6&89.4&66.9&73.8&62.0&89.5&83.2&87.6&95.8
&61.4&79.0&54.3&88.0&78.3&78.7 \\ \hline
Weak Sup.$^*$\!\cite{oquab2014weakly}
&96.7&88.8&92.0&87.4&64.7&91.1&87.4&94.4&74.9&89.2&76.3&93.7&95.2&91.1&97.6
&66.2&91.2&70.0&94.5&83.7&86.3 \\ \hline
HCP Alex$^*$\!\cite{hcp_pami}
&97.7&83.2&92.8&88.5&60.1&88.7&82.7&94.4&65.8&81.9&68.0&92.6&89.1&87.6&92.1
&58.0&86.6&55.5&92.5&77.6&81.8 \\ \hline
HCP VD$^*$\!\cite{hcp_pami}
&\textbf{99.1}&\textbf{92.8}&\textbf{97.4}&\textbf{94.4}&\textbf{79.9}&\textbf{93.6}
&\textbf{89.8}&\textbf{98.2}&78.2&\textbf{94.9}&79.8&\textbf{97.8}&\textbf{97.0}&93.8&96.4
&\textbf{74.3}&\textbf{94.7}&71.9&96.7&88.6&\textbf{90.5} \\ \hline
\textbf{EPD} CaffeNet
&96.2&84.9&90.7&87.1&61.8&89.9&83.4&92.1&71.1&77.8&73.4&89.6&88.1
&89.8&96.4&63.6&82.9&63.7&93.1&82.2&82.9 \\ \hline
\textbf{EPD} VD &99.0&90.7&95.5&93.7&78.9&93.2&88.6&97.3&\textbf{80.5}&91.3&\textbf{81.6}&96.0&96.1&\textbf{95.2}
&\textbf{97.9}&70.0&93.6&\textbf{72.3}&\textbf{97.5}&\textbf{89.0}&89.9 \\ \hline
\end{tabular}
\vspace{-0.1cm}
\end{table*}
\subsubsection{Number of detectors} 
\begin{figure}[t]
  \centering
  \includegraphics[width=.38\textwidth]{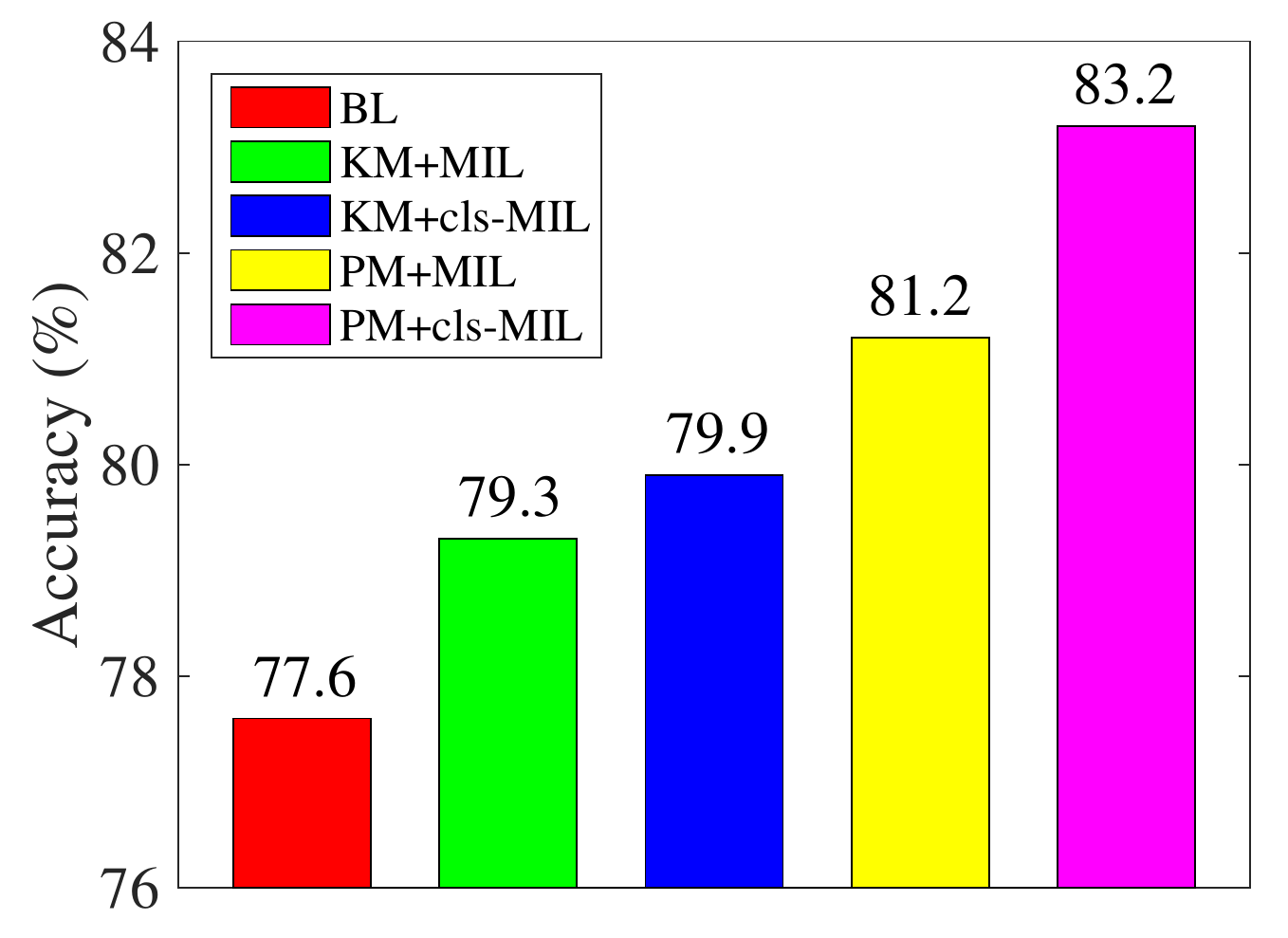} \\
 \vspace{-0.1cm}
  \caption{The classification performance comparisons with different configurations on Pascal VOC 2007 \emph{test} split. BL refers to baseline which max pooling CNN region features, KM is short for standard $k$-means pattern initialization algorithm, PM denotes the proposed pattern mining approach, MIL stands for standard multiple instance learning, and cls-MIL is the proposed confidence loss sparse MIL method. Results are based on model CaffeNet.}\label{ablation}
\vspace{-0.2cm}
\end{figure}

An advantage of the proposed approach is that detectors are trained from patterns with different coverage entropies. This enables us to greedily select detectors based on the entropy coverage criterion. As shown in Fig. \ref{ablation}, we add detectors orderly to probe how the number of detectors affect the classification performance. Note that the performance improves fast when a small number of detectors are used (\emph{e.g.}, from 1 to 10), it tends to be stable and even drops sightly when more detectors are added in. This is mainly because the subsequent detectors are not discriminative enough for classification. For computational efficiency, we fixed the number of detectors ($20$ per category for VOC 2007 and $30$ per category for MIT Indoor-67) for the following experiments.

\begin{figure}[t]
  \centering
  \includegraphics[width=.49\textwidth]{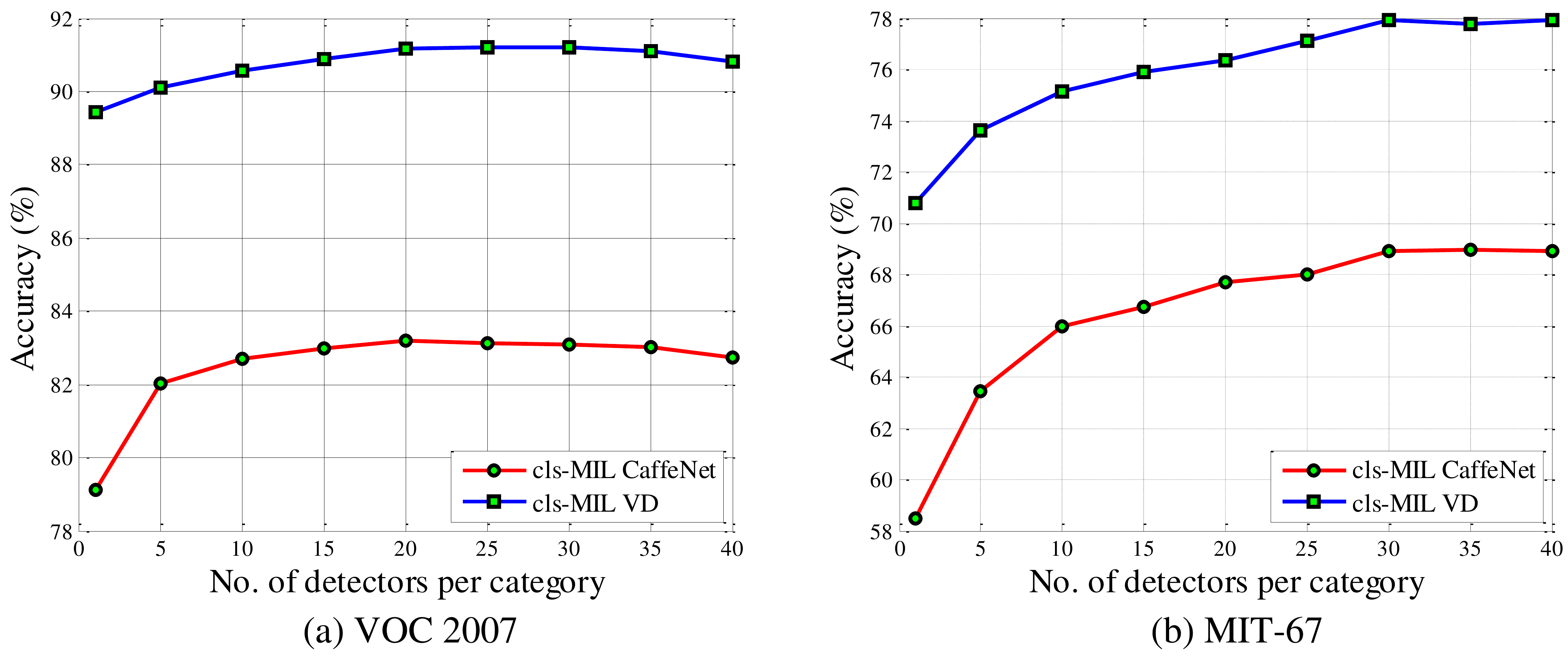} \\
 \vspace{-0.2cm}
  \caption{The classification performance with respect to the number of detectors per category on (a) VOC 2007 and (b) MIT Indoor-67. The detectors are greedily selected via the entropy coverage criterion. }\label{no_detectors}
\vspace{-0.1cm}
\end{figure}
\subsubsection{Number of region proposals} In order to probe the performance with respect to the number of candidate region proposals, we select the number of region proposals in different settings. Table \ref{no_regions} shows the results on VOC 2007 by varying the number of region proposals. The performance are relatively stable (from 2000 to 300 region proposals, only $1.3\%$ drop). Considering the performance and computational efficiency, we choose the number of region proposals as $500$.

\subsubsection{Effects of different modules} We now compare the results with different configurations to analysis how each module affect the final classification performance. Different modules are summarized as follows:

$\bullet$ BL: This is the baseline method which directly max pooling multiple region proposal features for classification. It is introduced to help understand how the proposed approach improve the discriminative power of the detectors.

$\bullet$ PM$\And$KM: PM denotes the proposed pattern mining method in Sec. III A, while KM is the standard $k$-means clustering method that is widely used for detector initialization in previous algorithms \cite{parizi2014automatic}, \cite{singh2012unsupervised}, \cite{sun2013learning}. For fair comparisons, we perform $k$-means clustering on the selected patches with the number of clusters setting as $20$.

$\bullet$ MIL$\And$cls-MIL: MIL stands for standard multiple instance learning method which mines new positive sample without considering the confidence of each bag, and cls-MIL is the confidence loss sparse MIL detector optimization strategy proposed in Sec. III B.

As shown in Fig. \ref{ablation}, both $k$-means and multiple instance learning do help to improve the classification performance, nevertheless with limited gains. The proposed pattern mining and cls-MIL method surpass the counterparts consistently, \emph{e.g.}, pattern mining improves the accuracy from $79.9\%$ ($k$-means) to $81.2\%$, and cls-MIL obtains an accuracy improvement of $2\%$ ($83.2\%$ vs $81.2\%$) comparing with standard MIL. We also find that detector initialization really counts for multiple instance learning, even for the modified cls-MIL ($79.9\%$ with $k$-means, and $83.2\%$ with pattern mining). This is widely discussed in previous approaches which aim to develop efficient pattern mining methods \cite{li2015mid}, \cite{bilen2015weakly} for detector initialization. However, few works emphasis detector optimization. We demonstrate that both modules are essential, and a combination of them achieves considerable performance improvement.
\subsection{Image Classification}
\begin{table}[t]
\centering
\caption{Comparisons of recognition performance on MIT Indoor-67. Clustered detectors refer to directly using clustered exemplar-SVM detector responses as features.} \label{rec_mit}
\vspace{-0.1cm}
\begin{tabular}{l|c|c}
\hline
Method & Dimension & Accuracy (\%) \\ \hline
DMS \cite{doersch2013mid}       & 13K  & 64.0 \\ \hline
DSFL \cite{parizi2014automatic} & 13K  & 77.1 \\ \hline
MOP-CNN \cite{gong2014multi}    & 13K & 68.9 \\ \hline
MDPM \cite{li2015mid}           & 3.3K & 77.6 \\ \thickhline
FC-CNN CaffeNet \cite{jia2014caffe} & 4K   & 60.3 \\ \hline
MR-CNN CaffeNet \cite{jia2014caffe} & 4K   & 65.1 \\ \hline
Clustered Detectors CaffeNet    & 2K   & 66.3 \\ \hline
\textbf{EPD} CaffeNet                    & 2K   & 69.0 \\ \hline
\textbf{EPD} VD                          & 2K   & 77.9 \\ \hline
\textbf{EPD} VD+\cite{Simonyan14c}       & 6K   & \textbf{80.1} \\ \hline
\end{tabular}
\vspace{-0.1cm}
\end{table}
\subsubsection{Object Recognition}
Table \ref{rec07_results} and \ref{rec12_results} show the object recognition results of the proposed approach on Pascal VOC 2007 and 2012 \emph{test} splits, respectively. In order to make fair comparisons, we extract CNN features from multiple region proposals, and max-pooling the region features into a final representation, which we refers to MR-CNN. Then the only difference between MR-CNN and our method is the detectors since they make use of the same region proposals. From Table \ref{rec07_results} we can see that the proposed detectors improve the classification performance considerably, achieving accuracies of $83.2\%$ with CaffeNet, and $91.3\%$ with very deep model, which bring $5.6\%$ and $2.2\%$ gains comparing with using CNN features.

There exist many previous approaches that report classification results on Pascal VOC dataset, and we compare our results with some most recent ones. Most of previous approaches that achieve high classification results are based on network fine tuning \cite{Bilen_2016_CVPR}, \cite{oquab2014learning}, \cite{hcp_pami}. Since network fine tuning is hard for multi-label images, previous works  \cite{oquab2014learning} rely on object annotations to find category specific patches. In \cite{hcp_pami}, the authors proposed a weakly supervised classification framework via two-steps of network fine tuning, while it makes use of additional training data, which is more demanding. Our result ($91.3\%$) is slightly better than the best performing one ($90.9\%$) \cite{hcp_pami}, demonstrating that the traditional optimization approaches are able to achieve competing results with CNN fine tuning. Furthermore, the proposed features are complementary with CNN features, and achieve an accuracy of $92.2\%$ when combined. For VOC 2012, our method obtains an accuracy of $89.9\%$, which is slightly worse than \cite{hcp_pami} ($90.5\%$) that makes use of additional training images. The reason lies in that CNN-based methods are powerful as the training data grow, while MIL-based methods are relatively robust to the amount of data.
\begin{figure}[t]
  \centering
  \includegraphics[width=0.49\textwidth]{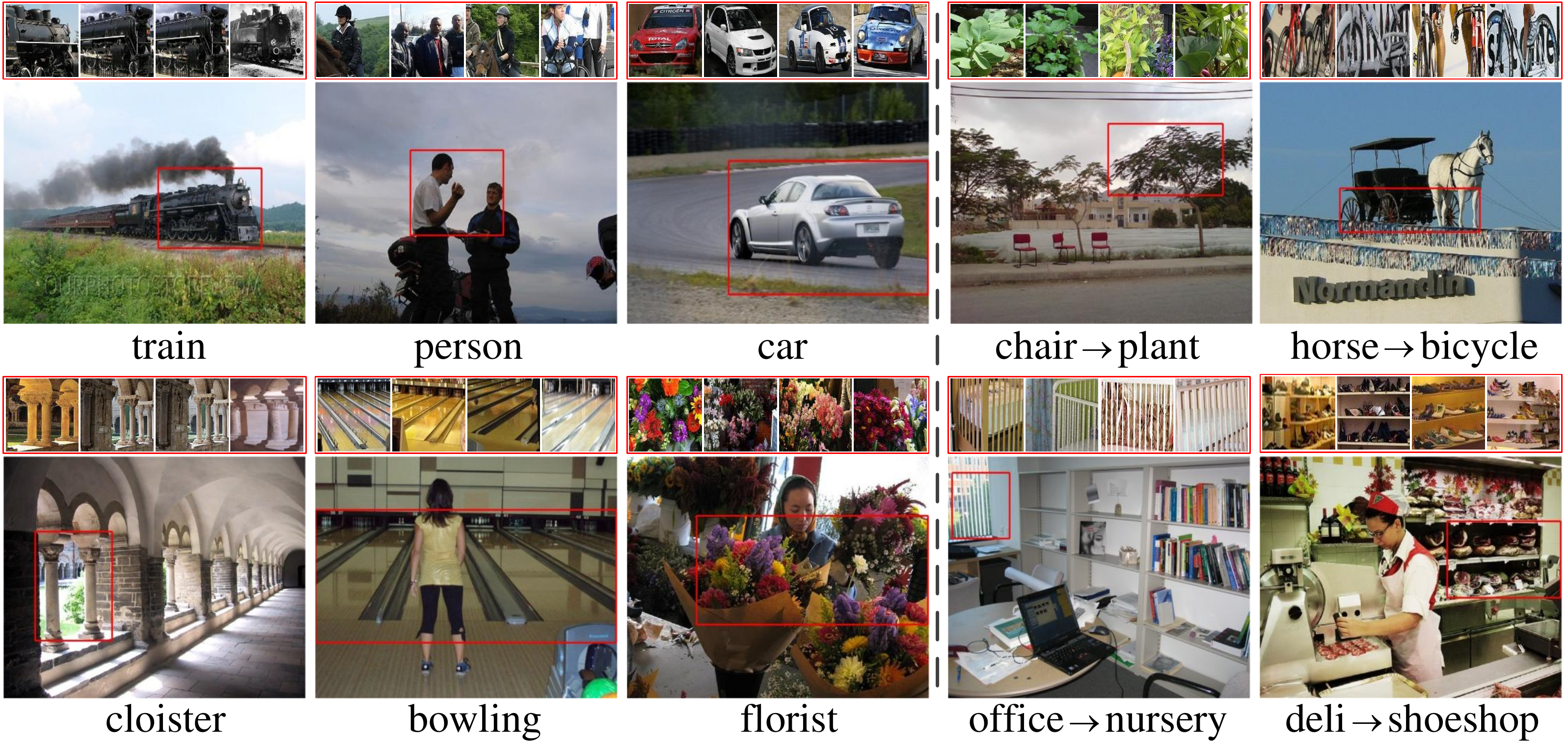} \\
  \vspace{-0.1cm}
  \caption{Some visualizations of the correct and incorrect classification. We show the top detection that makes it look like the corresponding category, and some patches that the detectors are trained on.}\label{vis_det_patch}
\vspace{-0.2cm}
\end{figure}
\subsubsection{Scene Recognition} Table \ref{rec_mit} compares the recognition results on MIT Indoor-67. MR-CNN denotes max-pooling multiple region features for representation, and FC-CNN refers to directly extract a single global feature from the whole image. Clustered detectors denote the method which relies on the responses of the clustered exemplar-SVM detectors as features. From Table \ref{rec_mit} we observe that:

$\bullet$ MR-CNN is much better than FC-CNN. Using CaffeNet model, the accuracy is $65.1\%$ with MR-CNN, and $60.3\%$ with FC-CNN. This demonstrate that local features are crucial for scene recognition.

$\bullet$ The features using clustered detector responses ($66.3\%$) is better than MR-CNN ($65.1\%$), even with half dimension (2K versus 4K). This is mainly because CNN is primarily trained from the object centric images, instead of the scene centric images. As a result, the weak exemplar-SVM detectors still outperform MR-CNN due to the data specific representation.

$\bullet$ The proposed EPD is much better than the features with clustered responses. Benefit from the detector optimization strategy, our method obtains an accuracy of $69.0\%$, which brings about a $2.7\%$ improvement comparing with the clustered responses. The performance is boosted to $77.9\%$ when switching to the very deep model. Another observation is that the proposed features are complementary with CNN features, and achieve an accuracy of $80.1\%$ when combined.

There are some approaches which also aim at learning discriminative part detectors for recognizing indoor scenes. The method of \cite{parizi2014automatic} integrates detector learning and classification by jointly training, and \cite{doersch2013mid} poses mid-level pattern discovery as discriminative mode seeking via developing an extension of the classic mean-shift algorithm to density ratio estimation. Our method is closely related to \cite{li2015mid}, which also makes use of CNN activations for pattern mining. Our method achieves better result comparing with the best performing method ($80.1\%$ vs $77.6\%$). There exists a majority of algorithms which employ multiple region pooling for final feature representation. A typical representation is MOP-CNN \cite{gong2014multi} which uses VLAD to encode CNN activations into bag of words representation, and achieves an accuracy of $68.9\%$, our results ($69\%$) is comparable with \cite{gong2014multi} using the same model, but with much lower dimension (2K vs 13K).

\subsubsection{Visualizing Mid-level Patterns}
As an illustration, Fig. \ref{vis_det_patch} shows some discovered patterns on VOC 2007 (top row) and MIT-67 (bottom row) \emph{test} splits. We show the highest activation region per image, which offers a clue indicating why it is classified as the corresponding category. Specifically, given a test image and the category label that the image is classified with (no matter correct or not), we employ category specific detectors to find which region responds most to the given category, and show some patches that the detector is trained on. For correctly classified images, there often exist discriminative patches that respond significantly to the corresponding detectors, \emph{e.g.}, on VOC 2007, the head of a train is important for recognizing the trains, and the upper body of a person is important for recognizing the persons. Similar results can be found on MIT-67, it is the pillar of a cloister that makes it look like a cloister, and the slide rail that makes bowling look like bowling. It is helpful to investigate why incorrect results happen, on VOC 2007, a classifier mis-classifies chair as the plant, or horse as bicycle, probably because there exist corresponding details, \emph{e.g.}, the wheel of the carriage is similar with bicycle wheels. Similar results can be found on MIT-67, the window of the office is misclassified as the bar of the baby bed, which is most discriminative for recognizing nursery. Actually, these details look similar, and it is hard to recognize them. However, these observations offer a direction to further improve the recognition performance.

\begin{table*}[t]
\caption{Object localization precision ($\%$) on VOC 2007 trainval images in terms of CorLoc metric.} \label{loc_results}
\scriptsize
\vspace{-0.1cm}
\setlength\tabcolsep{4.2pt}
\begin{tabular}{l|cccccccccccccccccccc|c}
\hline
method&aero&bike&bird&boat&bottle&bus&car&cat&chair&cow&
table&dog&horse&mbike&persn&plant&sheep&sofa&train&tv&mAP \\ \hline
Mimick \cite{li2016image}
&\textbf{73.1}&45.0&43.4&\textbf{27.7}&6.8&53.3&58.3&45.0&6.2&48.0&14.3&47.3&\textbf{69.4}&66.8&24.3&
12.8&51.5&25.5&\textbf{65.2}&16.8&40.0 \\ \hline
Con-Clust \cite{bilen2015weakly}
&66.4&59.3&42.7&20.4&21.3&63.4&\textbf{74.3}&\textbf{59.6}&21.1&\textbf{58.2}&14.0&38.5&49.5&60.0&19.8&
39.2&41.7&30.1&50.2&44.1&43.7 \\ \hline
MMIL \cite{cinbis2014multi}
&56.6&58.3&28.4&20.7&6.8&54.9&69.1&20.8&9.2&50.5&10.2&29.0&58.0&64.9&\textbf{36.7}&
18.7&56.5&13.2&54.9&59.4&38.8 \\ \hline
PLSA \cite{wang2014weakly}
&80.1&\textbf{63.9}&\textbf{51.5}&14.9&21.0&55.7&74.2&43.5&\textbf{26.2}&53.4&16.3&\textbf{56.7}&58.3&69.5
&14.1&38.3&58.8&47.2&49.1&\textbf{60.9}&\textbf{48.5} \\ \hline
\textbf{EPD} CaffeNet
&60.8&55.3&43.8&16.5&\textbf{29.4}&64.5&69.3&49.4&12.6&52.7&\textbf{29.7}&39.1&58.2&\textbf{81.1}&
34.0&\textbf{39.6}&58.8&\textbf{47.8}&59.3&53.1&47.7 \\ \hline
\textbf{EPD} VD
&60.8&58.8&40.8&17.6&24.8&\textbf{67.0}&68.1&50.0&12.2&48.6&27.4&36.5&58.2&78.7&
29.7&36.6&\textbf{63.9}&44.4&58.9&55.6&46.9 \\ \hline
\end{tabular}
\vspace{-0.1cm}
\end{table*}

\begin{figure*}[t]
  \centering
  \includegraphics[width=0.98\textwidth,height=4.35cm]{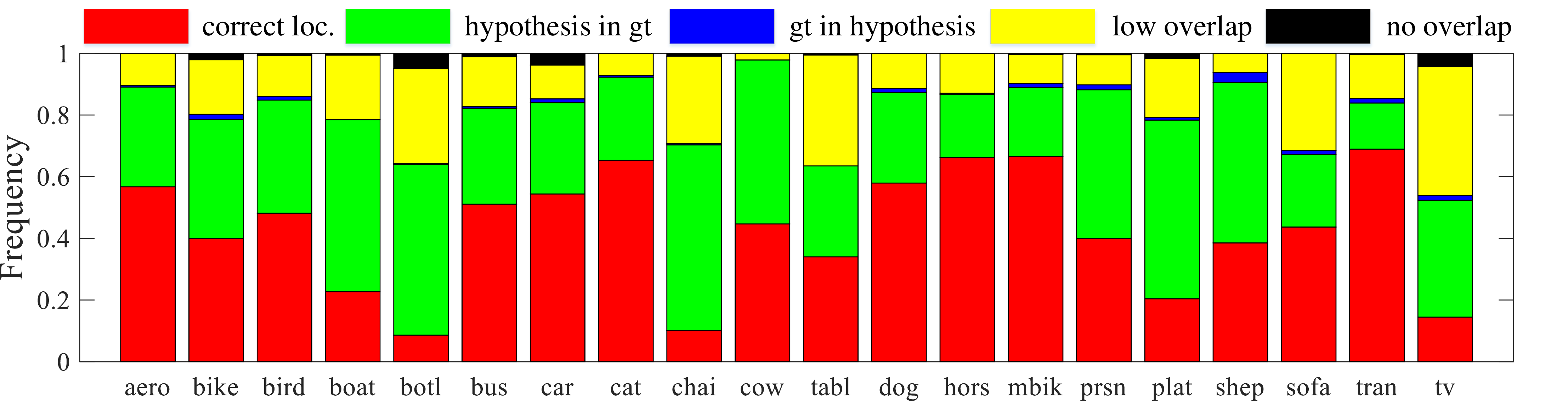} \\
  \vspace{-0.2cm}
  \caption{An illustration of the error distribution of the proposed localization method on Pascal VOC 2007 \emph{trainval} split.}\label{corloc}
\vspace{-0.1cm}
\end{figure*}
\begin{figure*}[t]
  \centering
  \includegraphics[width=0.95\textwidth]{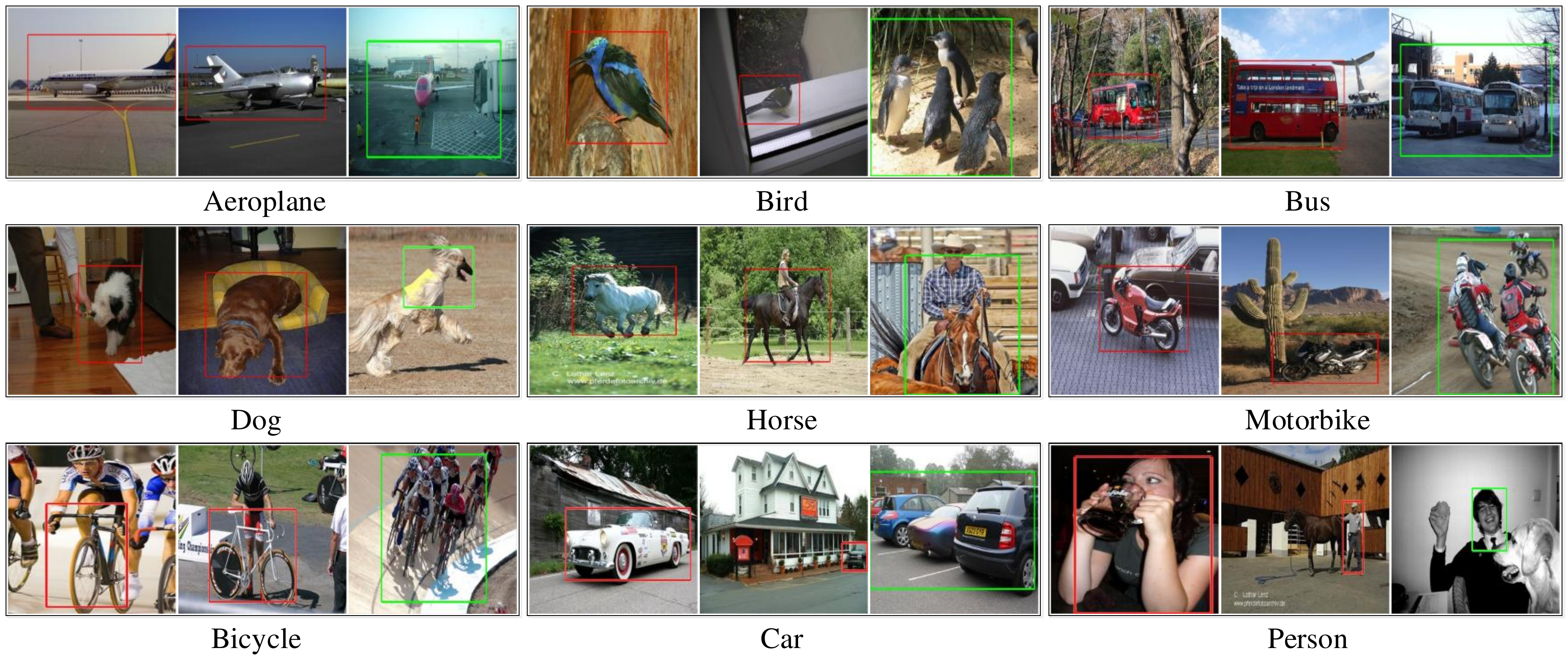}\\
  \vspace{-0.2cm}
  \caption{Examples of localization results on Pascal VOC 2007 \emph{trainval} split. The correct localization are marked with red bounding boxes, while the failed ones are marked with green. The failed results often come from localizing object parts or grouping multiple objects from the same class.} \label{detection}
\vspace{-0.2cm}
\end{figure*}
\subsection{Object Localization}
Table \ref{loc_results} shows the image localization results on Pascal VOC 2007 \emph{trainval} split. Benefit from the learned part detectors, the proposed localization strategy ($47.7\%$) is better than recent methods that is specifically designed for localization \cite{bilen2015weakly}, \cite{li2016image}, and is comparable with \cite{wang2014weakly} ($48.5\%$) which uses latent category learning for object localization. Another observation is that different from recognition, using deeper model does not bring about localization improvement (46.9\%). This can be explained with the fact that deeper models frequently focus on parts of the object instead of the whole object. Note that all these comparing methods are designed for localization, which often makes use of context information for better localization, while we rely on detectors which are learned for classification to uncover the connection between these two basic tasks. The results demonstrate that image classification and localization can be done simultaneously.
\subsubsection{Localization Error Analysis} In order to better understand the localization errors, following \cite{cinbis2014multi}, \cite{li2016image}, we summarize the errors to uncover the pros and cons of our localization method. Each predicted bounding box is categorized into the following five cases: 1) correct localization, IoU overlap is greater than $50\%$ with the ground truth. 2) hypothesis completely inside ground truth, 3) ground truth is completely inside the hypothesis, 4) no overlap, IoU equals to zero, and 5) low overlap, none of the above. Fig. \ref{corloc} shows the error distribution of the proposed method across $20$ categories on Pascal VOC 2007 \emph{trainval} set. It can be noted that among the failed modes, the most important failure modality of our method is that an object part is localized instead of the whole object. This is intuitive since in most situations, correct classification only demands catching local discriminative details.

\subsubsection{Visualizations and Limitations} Fig. \ref{detection} shows some localization results on Pascal VOC 2007 \emph{trainval} split. The correct localizations are marked with red bounding boxes, while the failed ones are marked with green. It can be shown that the proposed localization method is able to find objects where there is only one object from the same category, but is short of localizing multiple objects of the same category. Actually, it is the main challenge for weakly supervised localization \cite{cinbis2014multi}, and is a promising direction for future research.

\subsubsection{Classification versus Localization} Comparing classification (Table \ref{rec07_results}) with localization (Table \ref{loc_results}), we find that the least successfully recognized objects are \emph{bottle} ($79.1\%$) and \emph{chair} ($78.0\%$), which are also hard for localization ($24.8\%$ and $12.2\%$). This is because they usually occupy a small fraction of the image, and are within cluttered backgrounds. The exception is \emph{person}, which suffers a low localization accuracy ($29.7\%$), but with a high recognition accuracy ($98.7\%$). This can be explained by the fact that person is easy to be recognized by face, and usually, there exist multiple persons in an image, which offers abundant cues for recognition. In contrast, localization is failed when focusing on the face, and it is hard to tell apart individual person from the crowd.

\section{Conclusion}
In this paper, we propose an effective mid-level image representation approach for visual applications. The proposed framework aims at learning a collection of discriminative part detectors in a weakly supervised paradigm, which only needs the labels of training images, while does not need any object\,/\,part annotations. Our approach tackles several key issues in automatic part detector learning. First, we propose an efficient pattern mining technique via spectral clustering of exemplar-SVM detectors. Second, we formulate the detector learning as a confidence loss sparse MIL (cls-MIL) task, which considers the diversity of the positive instances, while avoid drifting away the well localized ones by assigning a confidence value to each positive instance. The proposed method shows notable performance improvements on several recognition benchmarks. Furthermore, we simultaneously considering classification and localization based on the learned detectors, and find that the accumulated responses of part detectors offer satisfactory localization performance, which bridges these two widely studied visual tasks.

{
\bibliographystyle{abbrv}
\bibliography{egbib}  
}
\end{document}